\documentclass[conference]{IEEEtran}
\IEEEoverridecommandlockouts
\usepackage{cite}
\usepackage{amsmath,amssymb,amsfonts}
\usepackage{graphicx}
\usepackage{textcomp}
\usepackage{xcolor}

\usepackage[utf8]{inputenc} % allow utf-8 input
\usepackage[T1]{fontenc}    % use 8-bit T1 fonts
\usepackage{hyperref}       % hyperlinks
\usepackage{url}            % simple URL typesetting
\usepackage{booktabs}       % professional-quality tables
\usepackage{amsfonts}       % blackboard math symbols
\usepackage{nicefrac}       % compact symbols for 1/2, etc.
\usepackage{microtype}      % microtypography
\usepackage{xcolor}         % colors
\usepackage{multirow}
\usepackage{makecell}
\usepackage{mathrsfs}
\usepackage{bm}
\usepackage{caption}
\usepackage{subcaption}
\usepackage{algorithm}
\usepackage{algorithmic}

\newcommand{\revise}[1]{{\color{black}#1}}

\def\BibTeX{{\rm B\kern-.05em{\sc i\kern-.025em b}\kern-.08em
    T\kern-.1667em\lower.7ex\hbox{E}\kern-.125emX}}
\begin{document}

\title{Relational Message Passing for Fully Inductive Knowledge Graph Completion
% \\
}

\author{\IEEEauthorblockN{Yuxia Geng}
\IEEEauthorblockA{\textit{College of Computer Science and Technology} \\
\textit{Zhejiang University}, Hangzhou, China \\
\textit{Donghai Laboratory}, Zhoushan, China
\\
gengyx@zju.edu.cn}
\and
\IEEEauthorblockN{Jiaoyan Chen}
\IEEEauthorblockA{\textit{Department of Computer Science} \\
\textit{The University of Manchester}\\
Manchester, United Kingdom \\
jiaoyan.chen@manchester.ac.uk}
\and
\IEEEauthorblockN{Jeff Z. Pan}
\IEEEauthorblockA{\textit{School of Informatics} \\
\textit{The University of Edinburgh}\\
Edinburgh, United Kingdom \\
j.z.pan@ed.ac.uk}
\and
\IEEEauthorblockN{Mingyang Chen}
\IEEEauthorblockA{\textit{College of Computer Science and Technology} \\
\textit{Zhejiang University}\\
Hangzhou, China \\
mingyangchen@zju.edu.cn}
\and
\IEEEauthorblockN{Song Jiang}
\IEEEauthorblockA{\textit{NAIE PDU} \\
\textit{Huawei Technologies Co., Ltd.}\\
Xi'an, China \\
jiangsong12@huawei.com}
\and
\IEEEauthorblockN{Wen Zhang\IEEEauthorrefmark{1}, Huajun Chen\IEEEauthorrefmark{1}\thanks{\IEEEauthorrefmark{1} Corresponding author.}}
\IEEEauthorblockA{\textit{Zhejiang University}, Hangzhou, China \\
\textit{Donghai Laboratory}, Zhoushan, China \\
\textit{AZFT Joint Lab for Knowledge Engine}\\
\{zhang.wen, huajunsir\}@zju.edu.cn}
}

\maketitle

\begin{abstract}
In knowledge graph completion (KGC), predicting triples involving emerging entities and/or relations, which are unseen when the KG embeddings are learned, has become a critical challenge.
Subgraph reasoning with message passing is a promising and popular solution. Some recent methods have achieved good performance, but they \textit{(i)} usually can only predict triples involving unseen entities alone, failing to address more realistic fully inductive situations with both unseen entities and unseen relations, and \textit{(ii)} often conduct message passing over the entities with the relation patterns not fully utilized.
In this study, we propose a new method named RMPI which uses a novel \textit{R}elational \textit{M}essage \textit{P}assing network for fully \textit{I}nductive KGC.
It passes messages directly between relations to make full use of the relation patterns for subgraph reasoning with new techniques on graph transformation, graph pruning, relation-aware neighborhood attention, addressing empty subgraphs, etc., and can utilize the relation semantics defined in the KG's ontological schema. 
Extensive evaluation on multiple benchmarks has shown the effectiveness of RMPI's techniques and its better performance compared with the existing methods that support fully inductive KGC.
RMPI is also comparable to the state-of-the-art partially inductive KGC methods with very promising results achieved.
Our codes, data and some supplementary experiment results are available at \url{https://github.com/zjukg/RMPI}.

\end{abstract}

\begin{IEEEkeywords}
Knowledge Graph, Inductive Knowledge Graph Completion, Message Passing, Link Prediction, Ontology
\end{IEEEkeywords}

\section{Introduction}
Knowledge Graphs (KGs)~\cite{PVGW2017} often suffer from incompleteness \cite{farber2018linked}.
Many KG completion (KGC) methods, including the schema aware ones~\cite{WPKD2020}, have been developed to discover missing facts (triples).  Many of them use KG embedding techniques,    encoding the KG entities and relations into a vector space with their semantics concerned, so that the missing facts can be inferred by computation on these vector representations (embeddings) \cite{wang2017knowledge,chen2020knowledge}.
However, these embedding-based methods often work in a \textit{transductive} setting, where the triples to predict involve only entities and relations that have already occurred in the embedding training triples. When some entities or relations are newly added during testing (a.k.a. \textit{unseen entities or relations}), they have to re-train the whole KG embeddings, which is not feasible in practice due to the quickly evolving nature and large sizes of many real-world KGs.

% , especially those representation learning based ones, have thus been proposed to improve KG completeness by discovering these missing triples, which is commonly referred as knowledge graph completion (KGC). 

\begin{figure}
\begin{subfigure}{.15\textwidth}
  \centering
  \includegraphics[width=1\linewidth]{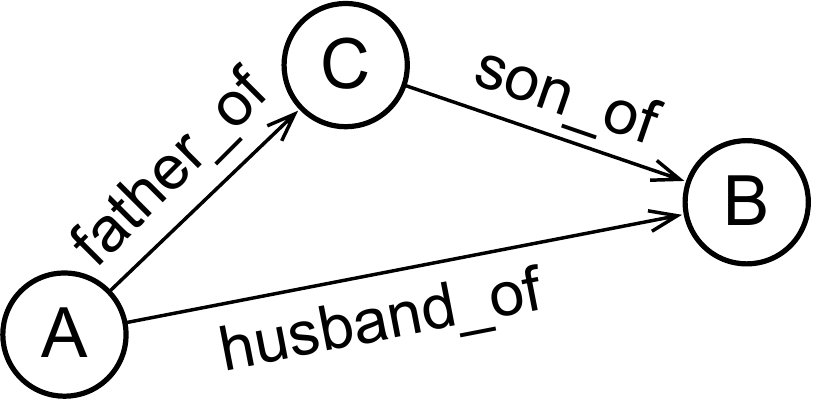}
  \caption{ }
  \label{fig:train_graph}
\end{subfigure}
\hfil
\begin{subfigure}{.15\textwidth}
  \centering
  \includegraphics[width=1\linewidth]{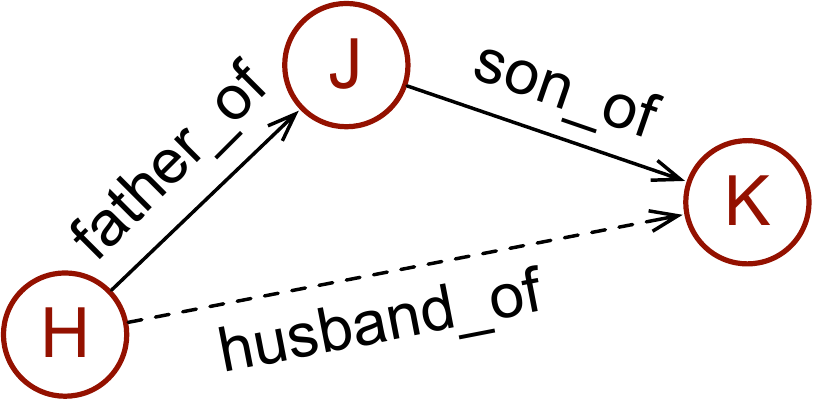}
  \caption{}
  \label{fig:partially_inductive_inference}
\end{subfigure}
\hfil
\begin{subfigure}{.15\textwidth}
  \centering
  \includegraphics[width=1\linewidth]{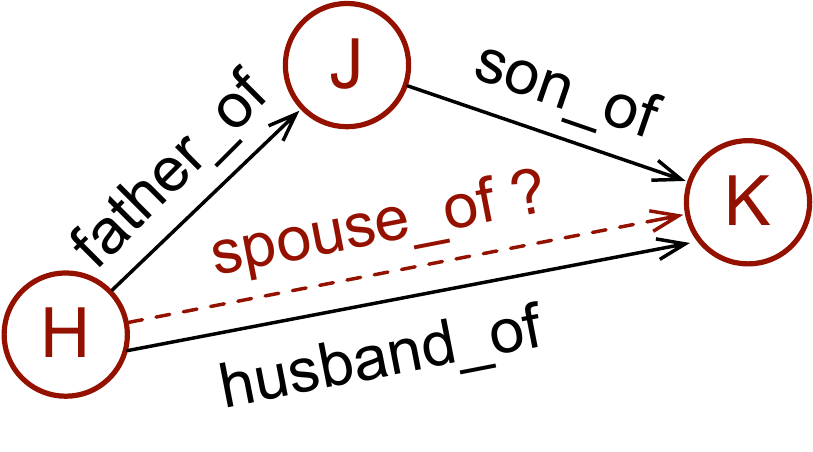}
  \caption{}
  \label{fig:fully_inductive_inference}
\end{subfigure}
\vspace{-0.1cm}
\caption{\small Examples on inductive KGC: (a) a training graph, i.e., a given KG whose embeddings have been learned; (b) a testing graph with unseen entities for \textit{partially inductive} completion; (c) a testing graph with both unseen entities and unseen relations (\textit{spouse\_of}) for \textit{fully inductive} completion.
% The nodes represent entities. 
The unseen elements are colored in red. 
%$H, J, K$ are three unseen entities newly added during inference.
}
\label{fig:example}
\end{figure}

%In consideration of the quickly evolving nature of KGs, where novel entities and relations are added frequently, repeatedly leaning the embeddings with new knowledge is not a practical solution.

Recently, there is an increasing number of  inductive KGC studies which aim to complete triples involving unseen entities or unseen relations without training the KG embeddings from the scratch.
%that are not seen during training (i.e., \textit{unseen entities}).
Among all these works, some try to obtain the embeddings of unseen entities or unseen relations using external resources (e.g., textual descriptions) \cite{shah2019open,xie2016representation,wang2021structure,qin2020generative,geng2021ontozsl} or auxiliary triples which associate unseen entities with seen entities \cite{hamaguchi2017knowledge,LAN,bhowmik2020explainable}.
Although these approaches can work, the additional resources they require are often   unavailable or have low quality, some even with   extra high computation costs, e.g. for text embedding. %, sometimes are relative high.
%on additional inputs are required, which are time-consuming and may not always be feasible.

An promising and widely investigated direction for addressing inductive KGC with unseen entities is to acquire high level semantics merely from the graph structure.
The relevant approaches often 
induce entity-independent logical rules that hold among the relations,
%and hence inherently generalizing to unseen entities, 
from the KG in either statistical \cite{meilicke2018fine} or end-to-end differentiable manners \cite{NeuralLP,DRUM,RuleN}.
For example, we can induce a rule $\textit{father\_of} (x,y) \wedge \textit{son\_of} (y,z)  \rightarrow \textit{husband\_of} (x,z)$ from the training graph in Fig. \ref{fig:train_graph}, and apply it to infer new triples in a testing graph with unseen entities, e.g., (\textit{H}, \textit{husband\_of}, \textit{K}) in Fig. \ref{fig:partially_inductive_inference}.
GraIL \cite{GraIL} is a typical and recent method of this kind, with the scalability and rule expressiveness improved by general graph topological features and a relation-aware graph neural network for message passing and learning over local subgraphs (see more details in Section~\ref{GraIl_formulation}).  
GraIL and its follow-up methods  \cite{TACT,CoMPILE} have shown the feasibility of using message passing over subgraphs for inductive KGC with promising results achieved on several benchmarks.
%in the context of KGs.

However, these methods mostly assume that all the relations in the testing stage are seen with embeddings learned, which is often violated in real-world evolving KGs,
\revise{especially those constructed via open information extraction systems e.g. NELL \cite{mitchell2018never} and those that are publicly editable e.g. Wikidata \cite{vrandevcic2014wikidata}.}
% \footnote{\url{http://rtw.ml.cmu.edu/rtw/}}
% \footnote{\url{https://www.wikidata.org/}}
Moreover, they often pass messages directly over entities with the patterns on relations not fully utilized.
Although some strategies have been developed to strengthen the role of relations, there is still much space to explore.
For convenience, we call the inductive KGC cases with only unseen entities during testing investigated in these works as \textbf{partially inductive KGC}, and call those more realistic and more challenging cases with both unseen entities and unseen relations during testing as \textbf{fully inductive KGC}\footnote{\revise{The term \textit{fully inductive} has been used in some inductive KGC works that only consider unseen entities \cite{GraIL,ali2021improving}, meaning the sets of entities seen during training and testing are disjoint, as against the other \textit{semi inductive} case where unseen entities have to be connected to trained seen entities.
In our paper, we name such a \textit{fully inductive} setting with only unseen entities as \textbf{partially inductive}, to distinguish it from the \textbf{fully inductive} setting we investigated.}}.
Fig. \ref{fig:example} provides clear examples on these two cases.
Currently, there have been few methods specifically developed for the \revise{latter}. The only one we know is MaKEr \cite{MaKEr} which also relies on the graph structure for prediction.

In this study, we aim to address fully inductive KGC with a method named RMPI, which includes a relation oriented message passing network for subgraph reasoning.
It first transforms a triple's surrounding subgraph in the original KG into a new relation view graph, where inter-relation features are more straightforwardly represented. It then learns the embedding of an unseen relation from the relational subgraph, by the relational message passing network, where novel graph pruning and neighborhood attention techniques are developed for both efficiency and effectiveness, with new neighborhood aggregations being used for addressing the issue of empty subgraphs.
As an example, for an unseen relation \textit{spouse\_of} in Fig. \ref{fig:fully_inductive_inference}, its embedding for predicting the triple with \textit{H} and \textit{K} can be obtained by aggregating the embeddings of neighboring relations \textit{husband\_of}, \textit{father\_of} and \textit{son\_of}.
%which have been well learned from the training graph.
Furthermore, RMPI allows the injection of the KG's ontological schema, which is quite common (see KGs like DBpedia \cite{auer2007dbpedia} and NELL \cite{mitchell2018never}) and contains complementary relation semantics, for more robust reasoning on fully inductive KGC. 
In summary, our main contributions are the following:
\begin{itemize}
    \item We are among the earliest to investigate fully inductive KGC, considering both subgraph structures and ontological schemas.
    \item We have proposed a robust KGC model named RMPI with novel techniques for effective relational message passing and subgraph reasoning, supporting both partially and fully inductive KGC.
    \item Extensive experiments have been conducted on $4$ newly-constructed benchmarks and $14$ public benchmarks from different KGs.
    RMPI and its variants often outperform the baselines including the state-of-the-art methods.
\end{itemize}

\section{Preliminary}\label{sec:pre}
We begin by first formally defining the task and notations, and then briefly introducing the background of subgraph-based inductive reasoning used in existing works.

\subsection{Problem Formulation}
% formulation of KGC
A KG is often denoted as $\mathcal{G} =\{ \mathcal{E}, \mathcal{R}, \mathcal{T}\}$, where $\mathcal{E}$ is a set of entities, $\mathcal{R}$ is a set of relations, and $\mathcal{T} = \{(h, r, t)| h, t \in \mathcal{E}; r \in \mathcal{R}\}$ is a set of relational facts in form of RDF\footnote{Resource Description Framework. See \url{https://www.w3.org/RDF/}.} triple.
$h$, $r$ and $t$ are called a triple's head entity (subject), relation (predicate) and tail entity (object), respectively.
KGC is then defined to predict an input candidate triple as true or not (i.e., triple classification), or predict the missing entity/relation in a triple with the other two elements given (i.e., entity/relation prediction), which is often achieved by filling the incomplete triple with a candidate entity/relation and feeding it into the model.
It is expected that the true (positive) triples are predicted with higher scores than those false (negative) ones~\cite{ARWP2021}.
In a commonly practiced partially inductive KGC setting, a set of unseen entities $\mathcal{E}'$ with $\mathcal{E} \cap \mathcal{E}' = \emptyset$ are newly emerged during prediction, the goal is then to predict a triple ($h,r,t$) with $r \in \mathcal{R}$ and $h,t \in \mathcal{E}'$.
$ \mathcal{G}$ is often known as the training graph, while the graph composed of the triples by the unseen entities $\mathcal{E}'$ and the relations $\mathcal{R}$ is often known as the testing graph.

In this study, we extend the above partially inductive KGC to fully inductive KGC by introducing a set of unseen relations $\mathcal{R}'$ with $\mathcal{R} \cap \mathcal{R}' = \emptyset$ to the testing graph.
Specially, we further investigate two situations:
one is a general testing graph involving both seen and unseen relations, while the other is a testing graph involving only unseen relations.
Formally, a triple given in the testing graph and a triple to predict can be denoted as ($h, r, t$) with $h, t \in \mathcal{E}'$, $r \in (\mathcal{R} \cup \mathcal{R}')$ or  $r \in \mathcal{R}'$.
We name the evaluation with the first testing graph as \textit{testing with semi unseen relations}, and name the evaluation with the second testing graph as \textit{testing with fully unseen relations}.

\begin{figure*}
  \centering
  \includegraphics[width=0.9\linewidth]{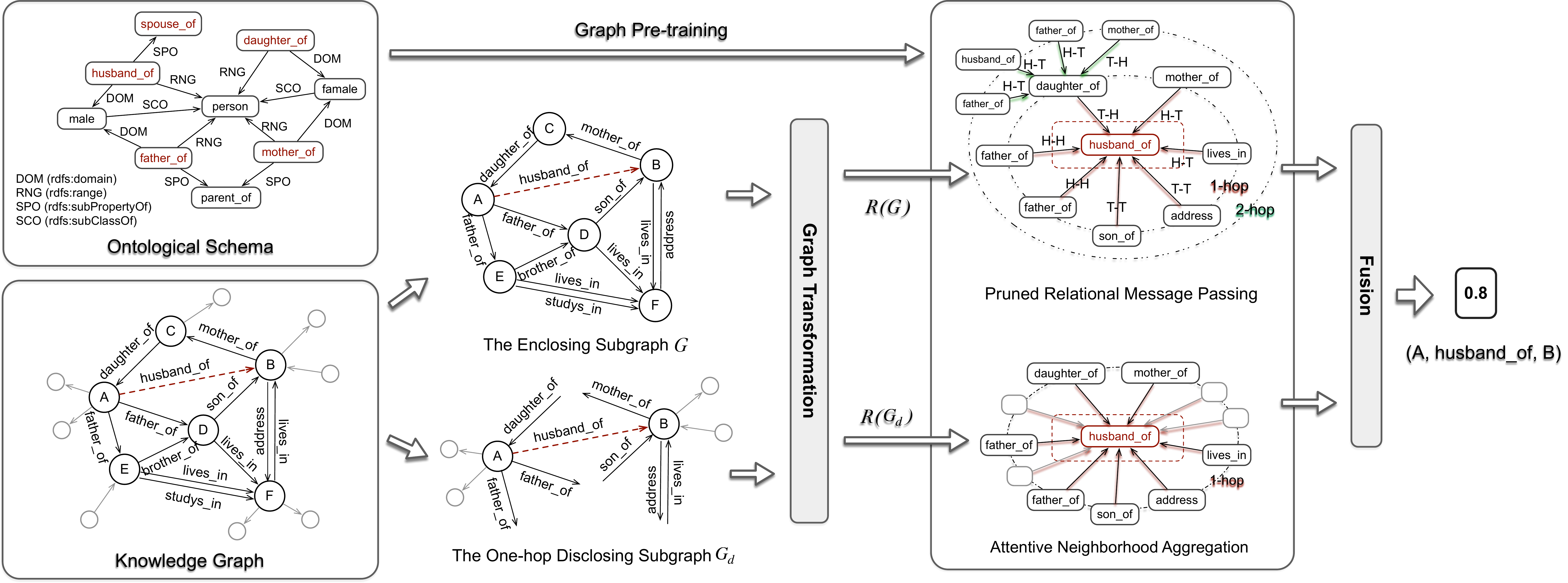}
\caption{The overall framework of our Relational Message Passing Network.}
\label{fig:framework}
\end{figure*}

\subsection{Subgraph-based Partially Inductive Reasoning}\label{GraIl_formulation}
To predict a target triple ($h, r, t$) where $r \in \mathcal{R}$, $h,t \in \mathcal{E}'$, and the embeddings of $h$ and $t$ are both not available, methods such as GraIL \cite{GraIL} learn and utilize the structural semantics from the subgraph around this triple in an entity-independent manner.
To better understand, we re-denote such a triple as ($u, r_t, v$) using the notations widely used in GNN, where $u$ and $v$ are referred as target head entity and target tail entity, respectively, and $r_t$ is the target relation.
The basic workflow of GraIL includes three steps.
It first extracts the $K$-hop enclosing subgraph %$\mathit{G}$ 
surrounding the target triple, denoted as $\mathit{G}_{(u,r_t,v)}$.
It then annotates the entities in the extracted subgraph according to their shortest distances to $u$ and $v$. Namely, for each entity $i$ in the subgraph, it is labeled with a tuple $(d(i, u), d(i, v))$, where $d(i, u)$ (resp.  $d(i,v)$) denotes the shortest distance between $i$ and $u$ (resp. $v$) without counting any path through $v$ (resp. $u$).
It finally summarizes the subgraph through a GNN encoder and scores the likelihood of the target triple using the encoded subgraph representation and the target relation's embedding.
The GNN-based encoder adopts the general message passing scheme where a node representation is iteratively updated by combining it with aggregation of its neighbors’ representations, and considers the multi-relation features of subgraph. Formally, the embedding of entity $i$ in the $k$-th GNN layer is given by:
\begin{align}
    \bm{h}_i^k &= ReLU(\sum_{r \in \mathcal{R}} \sum_{j \in \mathcal{N}_i^r} \alpha_{ij}^k \bm{W}_r^k \bm{h}_j^{k-1}  + \bm{W}_{self}^k \bm{h}_i^{k-1})
    \\
    \alpha_{ij}^k &= \sigma(\bm{A}_1^k \bm{s} + \bm{b}_1^k)
    \\
    \bm{s} &= ReLU(\bm{A}_2^k[\bm{h}_i^{k-1} \oplus \bm{h}_j^{k-1} \oplus \bm{r}_t^a \oplus \bm{r}^a] + \bm{b}_2^k )
\end{align}
where $\mathcal{R}$ is the set of relations in the KG, $\mathcal{N}_i^r$ denotes the neighbors of entity $i$ under relation $r$, $\bm{W}_r^k$ is the relation-specific transformation matrix used in the $k$-th layer, and $\bm{W}_{self}^k$ is the matrix for combining message from itself. $\alpha_{ij}^k$ denotes the attention weight at the $k$-th layer for the edge connecting $i$ and $j$ via relation $r$, which is computed by a function of the latent embeddings of $i$ and $j$ learned at previous layer and the learnable attention embeddings of $r$ and the target relation $r_t$.
The initial embedding $\bm{h}_i^0$ is represented using the node labels obtained in the second step (i.e., concatenating the one-hot vectors of $d(i, u)$ and $d(i, v)$), and the final embedding after $K$ layers $\bm{h}_i^K$ is output to generate the subgraph representation $\bm{h}_{\mathit{G}_{(u,r_t,v)}}^K$ and score the triple, as:
\begin{align}
    score(u, r_t, v) &= \bm{W}[\bm{h}_{\mathit{G}_{(u,r_t,v)}}^K \oplus \bm{h}_u^K \oplus \bm{h}_v^K \oplus \bm{r}_t]
    \\
    \bm{h}_{\mathit{G}_{(u,r_t,v)}}^K &= \frac{1}{|\mathcal{N}_{\mathit{G}}|} \sum_{i\in \mathcal{N}_{\mathit{G}}} \bm{h}_i^K
\end{align}
where $\mathcal{N}_{\mathit{G}}$ is the set of entities in the subgraph.
The subgraph is then represented by averaged pooling over all the output entity embeddings, and the likelihood of the target triple is scored through a linear layer with the subgraph representation, the embedding of the target relation $\bm{r}_t$, and the output embeddings of the target entity pair ($\bm{h}_u^K$ and $\bm{h}_v^K$), concatenated and fed.

The extracted enclosing subgraph are informative with some logical evidences (e.g., paths) that could help deduce the relation between two entities contained.
%e.g., some paths contain information that could imply the target relation.
While the distance-aware entity initial features represent an entity's relative positions w.r.t. the target triple's head and tail entities in the subgraph, thus capturing the structural semantics without the need of learning specific embeddings from the whole KG.
%While the distance-aware entity initial features avoid learning fixed embeddings for entities while only indicate their relative position in the graph to capture the structural semantics.
Finally, the applied GNN layers pass messages between entities iteratively to update the entity features, and the resultant representations are used to score the triple together with the learnable relation embeddings.
% \todo{measure the compatibility between the contextual subgraph and the target triple}.
% \todo{Also, the final triple scoring with four kinds of vectors can be well explained from the view of rule learning.}
% That is to say, for a logical rule in the form of $r_1(X, Z_1) \wedge r_2(Z_1, Z_2) \wedge ... \wedge r_k(Z_{k-1}, Y) \Rightarrow r_t(X,Y)$, where $r_t, r_1, ..., r_k$ are relations in the KG, $X, Z_1, ..., Z_{k-1}, Y$ are entity variables that can be instantiated by relevant entity constants, the left part is known as the \textit{rule body} while the right pair is the \textit{rule head}, the GraIL model with $k$ GNN layers can be used to parameterize the grounding of the rule body, and then the computed score is to indicate whether there is a true logical rule or not when grounding the rule head with $X=u$ and $Y=v$. 

\section{Methodology}\label{sec:meth}

\subsection{Overview}
In this paper, we make full use of the reasoning clues reflected by relations, and propose to pass messages (features) directly from relations to relations iteratively to infer the likelihood of a target triple, during which the entity features are not used, while the embedding of an unseen relation can be inferred and is optionally further augmented by relation semantics from the KG's ontological schemas.
To achieve this goal, we proposed a Relational Message Passing Network as shown in Fig. \ref{fig:framework}.
Briefly, given a target triple, it \textit{(i)} extracts an enclosing subgraph $\mathit{G}$ and transforms it to a new graph that can straightforwardly represent the relationships between relations via their co-occurrence in $\mathit{G}$, 
\textit{(ii)} propagates features between relations by multiple message passing layers  which are optimized with a graph pruning strategy for higher computation efficiency and a neighborhood attention mechanism for higher accuracy, 
and \textit{(iii)} computes the target triple's score, using the output representation of the target relation, during which the inductive embeddings of unseen relations are obtained through the same message passing layers, and they are further allowed to be enhanced by the ontological schema in an apriori manner.

The technical details of the first and the second steps are introduced in the following two subsections --- \textit{Subgraph Extraction and Transformation} and \textit{Relational Message Passing}, respectively, while the details of the third step are mostly covered by the following \textit{Reasoning with Unseen Relations} and \textit{Triple Scoring and Model Training}.
Considering the enclosing subgraph is sometimes empty where the relational message passing cannot be applied, we also propose to explore discriminative features from the target triple's disclosing subgraph for augmentation, which is introduced in the final subsection --- \textit{Dealing with Empty Subgraphs}.

\begin{figure}
\begin{subfigure}{.18\textwidth}
  \centering
  \includegraphics[width=1\linewidth]{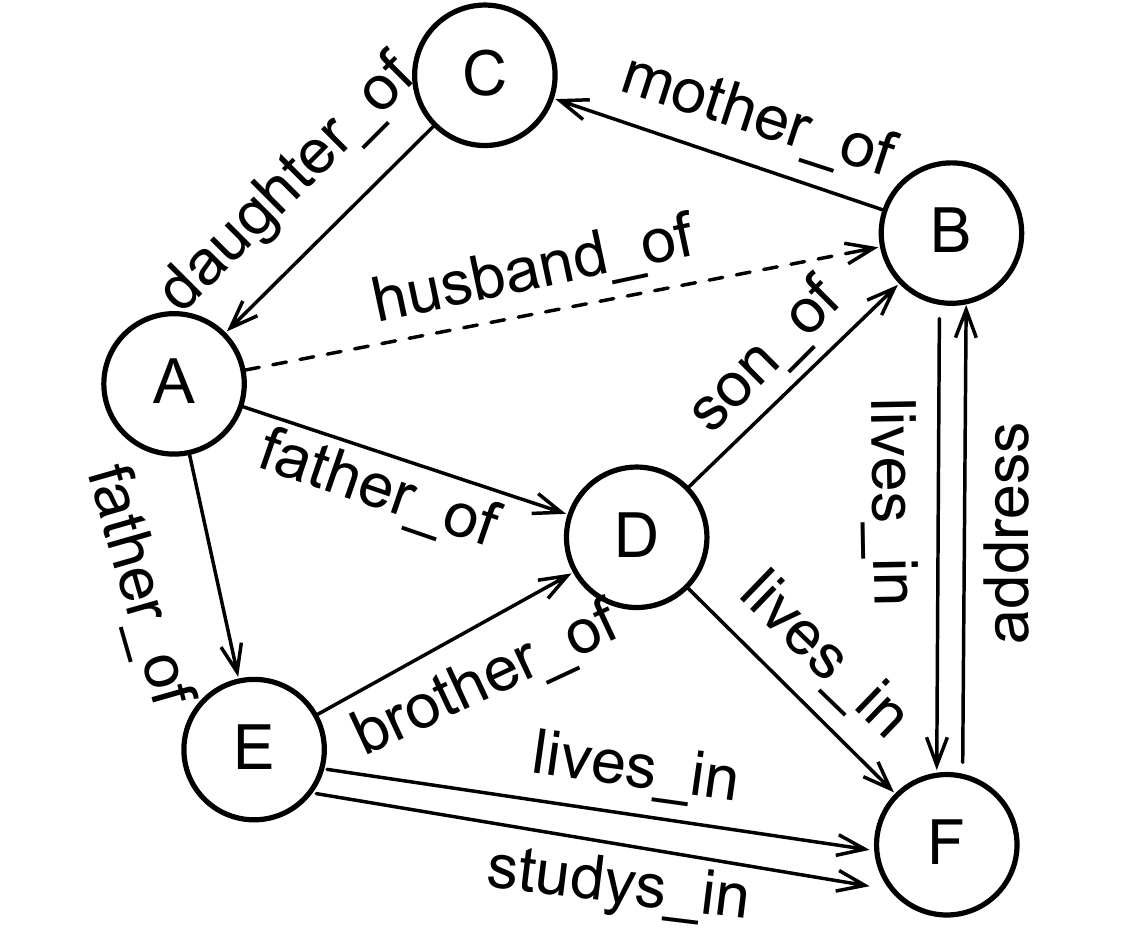}
  \caption{\small Graph in Entity View}
  \label{fig:entity_view_graph}
\end{subfigure}
\hfil
\begin{subfigure}{.26\textwidth}
  \centering
  \includegraphics[width=1\linewidth]{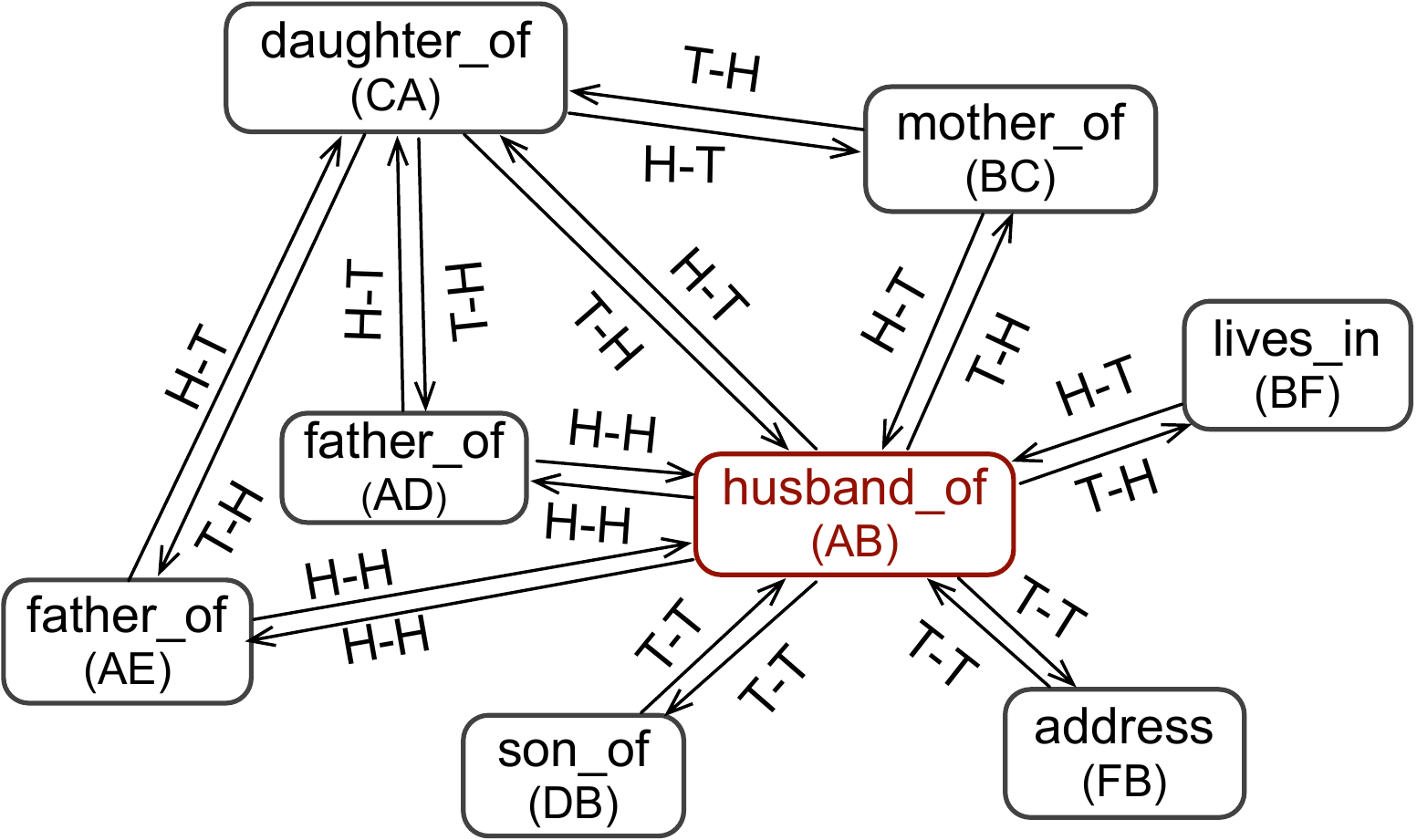}
  \caption{\small Graph in Relation View}
  \label{fig:relation_view_graph}
\end{subfigure}
\newline
\vspace*{0.2cm}
\newline
\begin{subfigure}{.5\textwidth}
  \centering
  \includegraphics[width=.7\linewidth]{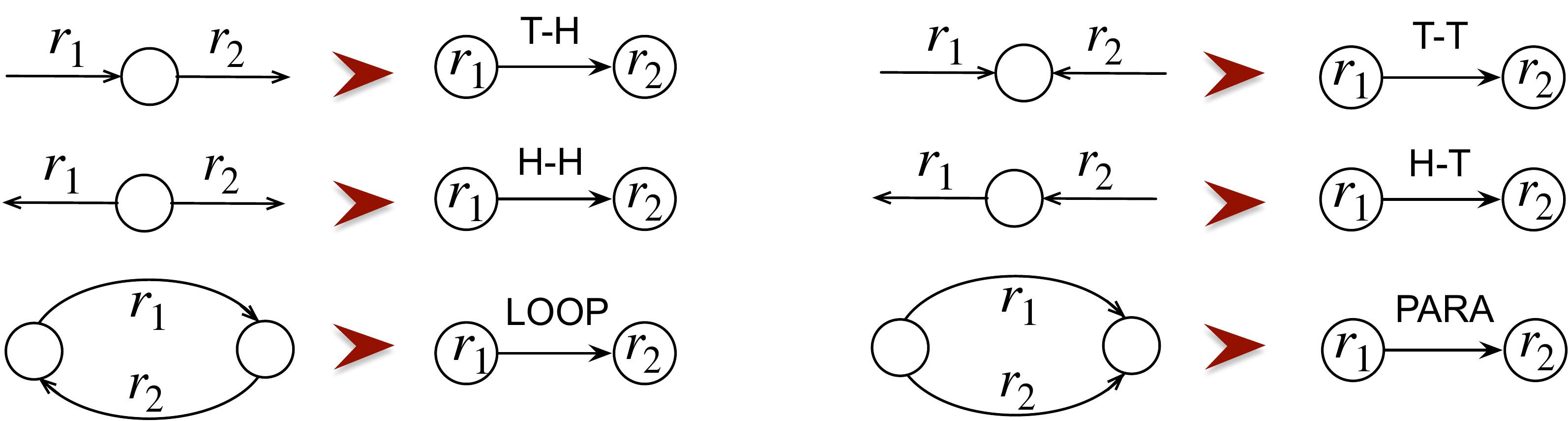}
  \caption{\small Connection Patterns of Relations
  }
  \label{fig:topo_pattern}
\end{subfigure}
\vspace{-0.1cm}
\caption{\small (a) and (b): A running example of transforming an entity-view subgraph to a relation-view subgraph (a fragment for the relation nodes \textit{husband\_of} and \textit{daughter\_of}).
(c): connection patterns (relationships) of relations according to their commonly associated entities.
}
\label{fig:graph_trans}
\end{figure}

\subsection{Subgraph Extraction and Transformation}\label{graph_transformation}
Given a target triple ($u, r_t, v$), its $K$-hop enclosing subgraph $\mathit{G}$ is extracted by the following steps.
First, the $K$-hop incoming and outgoing neighbors of the two target entities $u$ and $v$ from the original KG are collected, denoted as $\mathcal{N}_K(u)$ and $\mathcal{N}_K(v)$, respectively.
Then, the enclosing subgraph is generated by taking the intersection of the neighbor sets, i.e., $\mathcal{N}_K(u) \cap \mathcal{N}_K(v)$, pruning the nodes that are isolated or at a distance greater than $K$ from either $u$ or $v$, and adding the triples (i.e., edges) whose head and tail both belong to the pruned entity set.
In this way, we obtain a subgraph of distance at most $K+1$ from the target head to the target tail or vice versa.

$\mathit{G}$ is actually a graph in entity view as the original KG, where the graph nodes correspond to entities and the edges are labeled by relations.
%which is commonly accepted to present the knowledge graph.
To more straightforwardly represent the directed adjacencies between relations, we make a transformation following the idea of line graph \cite{harary1960some}, that is, a new graph in relation view is generated, during which all edges (identified by the labeled relation and its connected entity pairs) in the original graph are transformed into nodes, and every two new nodes are connected if and only if their corresponding relations are associated with at least one common entity in the original graph.
We denote such a relation-view subgraph of $\mathit{G}$ as $\mathit{R}(\mathit{G})$.
A running example is present in Fig. \ref{fig:graph_trans} (a) and (b).

Moreover, to represent different connection patterns (relationships) of the relations, we review the topological structure of the original entity-view graph and refer to \cite{TACT} and \cite{MaKEr} to define $6$ types of edges for the new relation-view graph, as shown in Fig. \ref{fig:topo_pattern}. 
For example, the edge \verb|H-H| connecting $r_1$ to $r_2$ is defined since the \underline{h}ead entity of relation $r_1$ is also the \underline{h}ead entity of relation $r_2$ in the original graph, \verb|T-H| means that the \underline{t}ail entity of $r_1$ is also the \underline{h}ead entity of $r_2$; \verb|PARA| indicates the head entities and the tail entities of $r_1$ and $r_2$ are all identical; while the crossed heads and tails are represented by \verb|LOOP|.
These types of edges model diverse relational connections for the graph in relation view.

\subsection{Relational Message Passing}
We next show how to perform message passing on $\mathit{R}(\mathit{G})$.
Since $\mathit{R}(\mathit{G})$ is much larger and denser than the original graph $\mathit{G}$, conducting message passing over all relation nodes is more computational inefficiency, as theoretically proven by Wang et al. \cite{PATHCON}.
Motivated by the neighborhood sampling strategy used in GraphSAGE \cite{GraphSAGE} --- a GNN model for large scale graphs, which first samples a subset of neighbors for each graph node (up to depth $K$) and then performs message passing on this sampled graph to generate its embedding instead of considering the whole neighborhood.
We also consider not updating all the graph nodes in one iteration.
Considering that our ultimate goal is to iteratively pass (aggregate) the information for the target relation $r_t$ from its $K$-hop neighbors, we propose to \textit{(i)} sample nodes on the graph according to whether they fall in the $K$-hop neighborhood of the target relation $r_t$ to generate a tree-like graph with the target relation as the root node, i.e., \textbf{a target relation-guided graph pruning strategy},
and \textit{(ii)} aggregate and update only the representations that are necessary to satisfy the recursion at each depth.

\begin{figure}
\renewcommand{\algorithmicrequire}{\textbf{Input:}}
\renewcommand{\algorithmicensure}{\textbf{Output:}}
\begin{algorithm}[H]
\caption{\small Relational Message Passing Algorithm}
\label{alg:algorithm_detail}
\begin{algorithmic}[1]
\renewcommand{\algorithmicrequire}{\textbf{Input:}}  % Use Input in the format of Algorithm
\renewcommand{\algorithmicensure}{\textbf{Output:}} % Use Output in the format of Algorithm
\small \REQUIRE Relational Subgraph $\mathit{R}(\mathit{G})$ with node set $\mathit{V}$ and edge set $\mathit{E}$; input features $\{\bm{h}_{r_i}^0, \forall r_i \in \mathit{V}\}$; depth (number of GNN layers) $K$; the neighbor set $\mathcal{N}_{r_t}$ of $r_t$ with $\mathcal{N}_{r_t}^0 =\{r_t\}$; differentiable aggregation and combination functions: $\text{AGGREGATE}^k$ and  $\text{COMBINE}^k$ for $k=1, 2, ..., K$
%$\forall k \in {1, ..., K}$
\small \ENSURE Propagated representations $\bm{h}_{r_t}^K$ for target relation $r_t$  %%output
\FOR {$k=1, 2, ..., K$}
\STATE get $r_t$'s $k$-hop incoming neighbors $\mathcal{N}_{r_t}^k=\{r_j  | (r_j, e, r_i) \in \mathit{R}(\mathit{G}); r_i \in \mathcal{N}_{r_t}^{k-1}; e \in \mathit{E}\}$
\ENDFOR
\FOR{$k=1, 2, ..., K$}
\STATE message passing for nodes $r_i \in \mathcal{N}_{r_t}^{\leq (K-k)}$:
\\
\STATE $\bm{h}_{\mathcal{N}_{r_i}}^k \leftarrow \text{AGGREGATE}^k(\{\bm{h}_{r_j}^{k-1}, \forall r_j \in \mathcal{N}_{r_i}\})$\;
\\
\STATE
$\bm{h}_{r_i}^k \leftarrow \text{COMBINE}^k(\bm{h}_{\mathcal{N}_{r_i}}^k, \bm{h}_{r_i}^{k-1})$
\ENDFOR
\RETURN $\bm{h}_{r_t}^K$
\end{algorithmic}
\end{algorithm}
\end{figure}

Algorithm \ref{alg:algorithm_detail} provides detailed pseudocode for illustrating the above procedure.
Specifically, given a message passing model with $K$ GNN layers, we first start from the target relation node and forward sample its full neighborhood sets up to depth $K$ in 1-3 steps.
In view of the direction of message passing, we sample all the incoming neighbors for each node.
Then, the message passing is conducted layer-by-layer in steps 4-8. In particular, at the $k$-th GNN layer with $k \in \{1, 2, ..., K\}$, we compute the latent representations for only neighbor nodes that are within the $(K-k)$ hops by aggregating the features of their directed neighbors (step 6) and combining the features of themselves to make an update (step 7), as these nodes will contribute their features to the representation learning of the target relation node in the future message passing.
For example, when $K=3$, the first layer computes latent features for nodes that are within $2$ hops; while the last layer only aggregates neighborhood features for the root node.
% Fig. \ref{fig:rmpi_propagation} provides a running example at each iteration step.
Finally, we take the embedding of the target relation output at the last layer, which have already fused the messages from its $K$-hop neighbors, to make a prediction. 
In this way, we extremely reduce the computation cost with a decreasing number of nodes that participate in the aggregation and are updated at each iteration.

The resulting message passing architecture gives the flexibility to plug in different AGGREGATE and COMBINE functions.
In consideration of the diverse connections between relation nodes illustrated by different edge types, we follow the idea of R-GCN \cite{RGCN} to model them in the aggregation.
In addition, we also apply \textbf{a target relation-aware neighborhood attention mechanism} to highlight the neighbors that are highly related to the target relation.
Formally, the AGGREGATE function at the $k$-th GNN layer is defined as follows:
\begin{align}
    \bm{h}_{\mathcal{N}_{r_i}}^k &= \sigma_1(\sum_{e=1}^6 \sum_{r_j\in \mathcal{N}_{r_i}^e} \alpha^k_{r_tr_j} \bm{W}_e^k \bm{h}_{r_j}^{k-1})
    \label{eq:aggregation_function}
    \\
    \alpha_{r_tr_j}^k &= \frac{exp(\sigma_2((\bm{h}_{r_t}^{k-1})^T \cdot \bm{h}_{r_j}^{k-1}))}{\sum_{r_{j'} \in \mathcal{N}_{r_i}^e}exp(\sigma_2((\bm{h}_{r_t}^{k-1})^T \cdot \bm{h}_{r_{j'}}^{k-1}))}
\end{align}
where $\mathcal{N}_{r_i}$ represents the directed incoming neighbors of node $r_i$, $\bm{h}_{\mathcal{N}_{r_i}}^k$ is the aggregated neighborhood vector;
$\mathcal{N}_{r_i}^e$ denotes the neighbors under edge type $e$ and $\bm{W}_e^k$ is the corresponding transformation matrix at the $k$-th layer;
$\bm{h}_{r_j}^{k-1}$ is the representation of a neighbor $r_j$ learned at previous iteration, while
$\alpha_{r_tr_j}^k$ is its attention weight with regard to the target relation $r_t$ in the $k$-th layer, the value is computed by the similarity of the representations of $r_j$ and $r_t$ learned at previous layer, following the assumption that when the neighbor is more related to the target relation, their representations are more similar.
A dot-product similarity is adopted here.
$\sigma_1$ and $\sigma_2$ are two non linear activation functions, they are ReLU and LeakyReLU (with negative input slope $\alpha=0.2$), respectively.

After aggregating the neighborhood features, we then combine $r_i$'s representation $\bm{h}_{r_i}^{k-1}$ obtained at previous layer with the aggregated vector to update its latent representation at the $k$-th layer. The combination is given by:
\begin{equation}
    \bm{h}_{r_i}^k =  \bm{h}_{\mathcal{N}_{r_i}}^{k-1} + \bm{h}_{r_i}^{k-1}
    \label{eq:combination_function}
\end{equation}

The initial representation of the node at the first layer, i.e., $\bm{h}_{r_i}^0$, is retrieved from the learnable relation embedding matrix that is randomly initialized.
In the last GNN layer, we perform an equal neighborhood aggregation for the target relation $r_t$ to preserve the directed neighborhood information and output its final representation by replacing \eqref{eq:aggregation_function} and \eqref{eq:combination_function} as:
\begin{equation}
    \bm{h}_{r_t}^K = ReLU(\sum_{e=1}^6 \sum_{r_j\in \mathcal{N}_{r_t}^e} \bm{W}_e^K \bm{h}_{r_j}^{K-1}) + \bm{h}_{r_t}^{K-1}
\end{equation}

\subsection{Reasoning with Unseen Relations}
\subsubsection{Utilizing the graph structure}
The above model successfully shows how message passes between relation nodes in the graph, where the features of a given relation node are gained from the representations of its neighboring relations.
In testing, for an unseen relation that first appears, we directly generate its embedding by aggregating the embeddings of its neighboring seen relations that have already been learned in training, using the well-trained aggregation functions previously defined, without modifying or re-training the model.

\subsubsection{Utilizing the ontological schema}\label{sec:onto}
A KG is often accompanied by an ontology as its schema for richer semantics and higher quality.
For example, the RDF Schema (RDFS)\footnote{\url{https://www.w3.org/TR/rdf- schema/}}-based ontological schema defines the types of entities (a.k.a. concepts), properties (including object properties i.e. relations, and data properties), concept and relation hierarchies, constraints (e.g., relation domain and range, and concept disjointness).
A number of vocabularies built in these ontologies bring in richer semantic relationships (connections) between seen and unseen relations.
Therefore, we also investigate utilizing the information from the ontology of a given KG for triple prediction with unseen relations, besides the graph structure.

\revise{Given the vocabulary set of RDFS~\cite{Pan2009}, we prefer those that are reasoning relevant and can be used to model correlations between KG relations, and finally select four types of vocabularies with separated relation semantics.} They are \textit{rdfs:subPropertyOf} for defining the subsumption relationship between relations, \textit{rdfs:domain} and \textit{rdfs:range} for defining the respective types of head and tail entities of relations, and \textit{rdfs:subClassOf} for defining the subsumption relationships between types.
All the semantics reflected by them can be represented using RDF triples.
For example, ($r_1, \textit{rdfs:subPropertyOf}, r_2$) reveals that $r_1$ is a child relation of $r_2$.
These triples constitute a schema graph whose nodes are relations or entity types and edges are vocabularies.
A segment can be found in the top left of Fig. \ref{fig:framework}. 

Next, we inject these semantics into our relational message passing network in an apriori manner.
Specifically, we first pre-train the schema graph which already covers all the target seen and unseen relations using KG embedding techniques e.g., the method by TransE \cite{transe}, which can learn vector representations for either seen or unseen relations with their semantic relationships kept in the vector space.
Then, we map these relation vectors to participate in the relational message passing to predict the target triple.
That is, the projected vectors are taken as the initial representations of nodes in $\mathit{R}(\mathit{G})$, i.e., $\{\bm{h}_{r_i}^0, \forall r_i \in \mathit{V}\}$.
The mapping function is implemented by two fully-connected linear layers, as
\begin{equation}
    \bm{h}_{r_i}^0 = \bm{W}_1(\bm{W}_2 \bm{h}_{r_i}^{onto})
\end{equation}
where $\bm{h}_{r_i}^{onto}$ is the relation vector of $r_i$ learned from the schema graph.
During testing, the semantic vectors of unseen relations are also projected to make a more robust subgraph reasoning together with those of seen relations.

\subsection{Triple Scoring and Model Training}
Finally, to obtain the score for the likelihood of the target triple ($u, r_t, v$), we pass the output representation of $r_t$ after $K$-layers message passing through a linear layer:
\begin{equation}
    score(u, r_t, v) = \bm{W} \bm{h}_{r_t}^K
    \label{eq:score_function}
\end{equation}

Following previous works, the whole model is trained by contrasting the scores of positive and negative triples using e.g. a margin-based ranking loss.
In general, each existing triple in the given KGs is taken as a positive triple, while a negative triples is generated by replacing its head (or tail) with a uniformly sampled random entity.
Accordingly, another enclosing subgraph is extracted for the negative target triple.
The loss function is as:
\begin{equation}
    \mathcal{L} = \sum_{i=1}^{|\mathcal{T}|} max(0, score(n_i)-score(p_i) + \gamma)
\end{equation}
where $\mathcal{T}$ is the set of all triples in the training graph; $p_i$ and $n_i$ denote the positive and negative triples, respectively; $\gamma$ is the margin hyperparameter, which is often a value greater than $0$ to score positive triples higher than the negative ones. 
During prediction, for a testing triple, the same enclosing subgraph is extracted and used to estimate its plausibility.

\subsection{Dealing with Empty Subgraphs}\label{method_NE}
In practice, quite a few triples, especially those randomly sampled negative triples, have empty enclosing subgraphs, i.e., no valid edge exists in the subgraph under hop $K$.
In this case, it becomes almost impossible to capture graph structure semantics to either infer the relation between a pair of entity or distinguish the positive triples from the negative ones.

Targeting this, we try to explore additional inputs from the $K$-hop disclosing subgraph of a target triple.
Specifically, we first take the union of the neighbor sets $\mathcal{N}_K(u)$ and $\mathcal{N}_K(v)$ of the target head entity $u$ and tail entity $v$, i.e., $\mathcal{N}_K(u) \cup \mathcal{N}_K(v)$, to generate the $K$-hop disclosing subgraph, then convert it to a relation-view subgraph following the same transformation step stated in Section \ref{graph_transformation}, and finally perform message passing on it to learn supplemental discriminative features for the target triple.
In view of the larger size of disclosing subgraph compared with the corresponding enclosing subgraph, we sample the one-hop neighbors of the target relation node and aggregate their features with an attention mechanism which assigns different weights to different neighbors for their importance towards the central target relation node.
The aggregation function is given by:
\begin{align}
    \bm{h}_{\mathcal{N}_{r_t}^d}^d &= \sigma_1 (\sum_{r_i\in\mathcal{N}_{r_t}^d} \alpha_{r_ir_t}^d \bm{W}^d \bm{h}_{r_i}^0)
    \label{eq:aggregation_function2}
    \\
    \alpha_{r_ir_t}^d &= \frac{exp(\sigma_2((\bm{W}^d \bm{h}_{r_t}^0)^T \cdot (\bm{W}^d \bm{h}_{r_i}^0)))}{\sum_{r_{i'} \in \mathcal{N}_{r_t}^d}exp(\sigma_2((\bm{W}^d \bm{h}_{r_t}^0)^T \cdot (\bm{W}^d \bm{h}_{r_{i'}}^0)))}
\end{align}
where $\mathcal{N}_{r_t}^d$ is the disclosing one-hop neighbor set of the target relation $r_t$, $\bm{h}_{r_i}^0$ is the embedding of a neighbor relation $r_i$, $\bm{W}^d$ is a linear transformation matrix applied to every node.
Dot-product similarity is adopted to compute the attention value. The activation functions $\sigma_1$ and $\sigma_2$ are set to ReLU and LeakyReLU, respectively.

Finally, we integrate the aggregated vector with the representation output from the enclosing subgraph to make more comprehensive triple scoring. Namely, \eqref{eq:score_function} is replaced by either
\begin{equation}
    score(u, r_t, v) = \bm{W} (\bm{h}_{r_t}^K
     + \bm{h}_{\mathcal{N}_{r_t}^d}^d) \label{eq:sum_combine}
\end{equation}
with a summation-based fusion function,
or 
\begin{equation}
    score(u, r_t, v) = \bm{W} (\bm{W}_3[\bm{h}_{r_t}^K
    \oplus \bm{h}_{\mathcal{N}_{r_t}^d}^d])
    \label{eq:conc_combine}
\end{equation}
with a concatenation-based fusion function by the linear transformation matrix $\bm{W}_3$.

\section{Evaluation}\label{sec:eva}
In our experiments, we first validate that our proposed framework is capable of handling fully inductive KGC utilizing either the graph structure alone or both the graph structure and the ontological schema.
Then, we show our proposed relational message passing has superiority in the case of partially inductive KGC with only unseen entities in comparison to existing subgraph reasoning methods.
Finally, we perform ablation studies and case studies for more detailed model analyses.
We begin by setting up the background for evaluation.

\subsection{Datasets and Ontologies}
The research on partially inductive KGC with unseen entities is evaluated with a series of benchmarks raised in GraIL \cite{GraIL}, which are derived from three famous transductive KGC datasets: WN18RR \cite{WN18RR}, FB15k-237 \cite{FB15k237} and NELL-995 \cite{NELL995}.
For robust evaluation, four versions of inductive benchmarks are extracted from each dataset with different sizes. Each benchmark consists of a training graph and a testing graph with disjoint entity sets.
In the training graph, $80\%$ of the triples are used to train the model, $10\%$ of them are used as the validation set; while in the testing graph, $10\%$ of the triples are taken as the final targets to predict.
The subgraph for each training, validation or testing triple is then extracted from the corresponding training and testing graphs.
The statistics are listed in Table \ref{dataset_GraIL}.

To support the evaluation of fully inductive KGC, where there exist unseen relations in the testing graph, we develop new benchmarks by re-combining the above $12$ inductive benchmarks.
Specifically, for an original benchmark, we hold the training graph but replace the testing graph by another benchmark whose testing graph has more relations.
For example, for the second benchmark from NELL-995 (NELL-995.v2 in short), which includes $88$ relations, we combine the testing graph of NELL-995.v3 having $122$ relations in total, $51$ out of which are not contained in NELL-995.v2, to generate a new benchmark named as NELL-995.v2.v3.
These re-combined datasets are denoted using the pattern ``XXX.v$i$.v$j$'', where XXX is the source transductive dataset, $i$ is the index indicating which version of inductive benchmark the training graph comes from, while $j$ is the version indicator for the source of the testing graph.
Obviously, for each dataset, we filter the testing graph to ensure that all its entities are not present in the corresponding training graph.
Since there are no enough unseen relations or testing triples under some combinations, we finally generate $4$ new datasets for evaluation, are shown in Table \ref{dataset_ours}.
Notably, the datasets constructed through the above steps actually fall in the setting of \textit{testing with semi unseen relations}, i.e., a portion of relations in the testing graph have been seen in the training graph while the rest are unseen.
Thus, we further extract subsets from the testing graphs by filtering the triples involving seen relations for the setting of \textit{testing with fully unseen relations}, which are entirely new graphs with only unseen entities and unseen relations.
As a result, for each dataset, we obtain two kinds of testing graphs labeled by `TE (semi)'' and ``TE (fully)'', respectively, as Table \ref{dataset_ours} shows.

\begin{table}
\centering
\caption{\small Statistics of the benchmarks. ``TR'' and `TE'' are short for the training graph and testing graph, respectively. ``\#R/E/T'' denotes the number of relation/entities/triples. }
\vspace{-0.1cm}
\begin{subtable}{1\linewidth}
\centering
\caption{\small The benchmarks for partially inductive KGC with unseen entities.}
\label{dataset_GraIL}
\vspace{-0.1cm}
% \resizebox{\linewidth}{!}{
\begin{tabular}{p{0.1cm}<{\centering}p{0.3cm}<{\centering}|p{0.2cm}<{\centering}p{0.4cm}<{\centering}p{0.6cm}<{\centering}|p{0.2cm}<{\centering}p{0.4cm}<{\centering}p{0.6cm}<{\centering}|p{0.2cm}<{\centering}p{0.4cm}<{\centering}p{0.6cm}<{\centering}}
\hline
&       & \multicolumn{3}{c|}{WN18RR} & \multicolumn{3}{c|}{FB15k-237}        & \multicolumn{3}{c}{NELL-995}         \\
&  & \#R & \#E & \#T & \#R & \#E & \#T & \#R & \#E & \#T \\
\hline
\multirow{2}{*}{v1} & TR 
& 9 & 2746 & 6678           
& 180 & 1594 & 5226
& 14 & 3103  & 5540 
\\
& TE  
& 8 & 922 & 1991
& 142 & 1093  & 2404
& 14 & 225 &1034
\\
% \midrule
\hline
\multirow{2}{*}{v2} & TR  
& 10 &6954 & 18968
& 200 & 2608 & 12085
& 88 & 2564 & 10109
\\
 & TE 
& 10 & 2757 & 4863
& 172 & 1660 & 5092
& 79 & 2086 & 5521         \\
% \midrule
\hline
\multirow{2}{*}{v3} & TR 
& 11 & 12078 & 32150
& 215 & 3668 & 22394
& 142 & 4647 & 20117
\\
& TE     
& 11 & 5084 & 7470
& 183 & 2501 & 9137
& 122 & 3566 & 9668
\\% \midrule
\hline
\multirow{2}{*}{v4} & TR 
& 9 & 3861 & 9842
& 219 & 4707 & 33916
& 76 & 2092 & 9289
\\
& TE  
& 9 & 7084 & 15157
& 200 & 3051 & 14554
& 61 & 2795 & 8520
\\% \bottomrule
\hline
\end{tabular}
\end{subtable}
\newline
\vspace*{0.2cm}
\newline
\begin{subtable}{1\linewidth}
\centering
\caption{\small The benchmarks for fully inductive KGC raised by us. The numbers in the brackets are the numbers of unseen relations.}
\label{dataset_ours}
\vspace{-0.1cm}
\begin{tabular}{c|ccc|ccc}
\hline
& \multicolumn{3}{c|}{NELL-995.v1.v3}
 & \multicolumn{3}{c}{NELL-995.v2.v3}        \\
 & \#R & \#E & \#T 
 & \#R & \#E & \#T
 \\
\hline
TR
& 14 & 3103 & 5540
& 88 & 2564 & 10109           
\\
TE (semi)
& 106 (98) & 2271 & 5550
& 116 (49) & 2803 & 6749
\\
TE (fully)
& \ 98 (98)  & 2246 & 5500
& \ 49 (49) & 1553   & 4174
\\
\midrule
& \multicolumn{3}{c|}{NELL-995.v4.v3}        & \multicolumn{3}{c}{FB15k-239.v1.v4}         \\
 & \#R & \#E & \#T
 & \#R & \#E & \#T
 \\
\hline
 TR 
& 76 & 2092 & 9289
& 180 & 1594 & 5226 
\\
TE (semi)
& 110 (53) & 3140 & 8308
& 200 (26) & 3051  & 14554
\\
TE (fully)
& \ 53 (53) & 2098  & 4739
& \ 26 (26) & 676 & 756
\\
\hline
\end{tabular}
\end{subtable}%
\end{table}

To better address the unseen relations, we also utilize KGs' ontologies to introduce richer relation semantics.
\revise{
%Regarding the four kinds of RDFS semantics introduced in Section~\ref{sec:onto}, 
For some KGs that have public ontologies, we can generate relation-aware schema graphs by extracting and re-organizing semantics from these ontologies; while for some KGs that do not have ontologies, it is possible to  construct the schema graph from scratch, for example, by inviting experts to set up the relation hierarchy and associate the domain and range constraints of relations. In our 
work, we experiment with NELL-995 and adopt the schema graph released in \cite{geng2022benchmarking}, which contains $1186$ nodes and $3055$ triples, covering all the relations and their semantics, and leave the evaluation of the ontology equipped setting on WN18RR and FB15k-237, which have no existing ontologies, as future works.}
TransE \cite{transe} is trained on the schema graph to generate meaningful semantic vectors for both seen and unseen relations.

\subsection{Evaluation and Implementation Details}
For more comprehensive evaluation, we tested completion tasks of triple classification and entity prediction.
Triple classification predicts
an input triple as true or not with a widely used binary classification metric AUC-PR (i.e., area under the precision-recall curve) reported, during which one negative triple is sampled for each triple in the test set that is often positive.
While for an incomplete triple ($h, r, ?$) or ($?, r, t$), the goal of entity prediction is to rank a set of candidate entities according to their predicted scores of being the tail (or head) entity, and see the rank of the ground truth — the smaller rank, the better performance.
Accordingly, the performance is evaluated by widely used metrics of Mean Reciprocal Ranking (MRR) and Hits@$n$ (i.e., the ratio of testing triples whose ground truths are ranked within top-$n$ positions) \cite{wang2017knowledge}.
For partially inductive KGC, we follow previous works to report Hits@10, while for fully inductive KGC, we make a comprehensive report with Hits@10 and MRR.
The negative triples in triple classification and entity prediction are both obtained by replacing the head (or tail) with a random entity.
Different from the one-to-one comparisons in triple classification, we follow previous works to rank the ground truth against another $49$ randomly sampled candidates in entity prediction.
We run each experiment $5$ times and report the mean results for a robust comparison.

We implement our model with PyTorch and use Adam as optimizer with learning rate of $0.001$, batch size of $16$ and margin value of $10$, which are the best configurations w.r.t the validation set.
For each triple, we sample $2$-hop enclosing and disclosing subgraphs, and apply two message passing layers on the enclosing subgraph.
For each relation, its embedding size is set to $32$, and the dimension of its semantic vector learned from schema graphs is  $300$.
Also, we perform an edge dropout with rate of $0.5$.
For more details, please refer to our codes. 

\begin{table*}[htbp]
\centering
\caption{Results of fully inductive KGC in the \textit{testing with semi unseen relations} setting. The best results are marked in bold.}\label{results_of_semi_unseen}
\vspace{-0.1cm}
\begin{subtable}{1\linewidth}
\centering
\caption{\textit{Random Initialized}: randomly initializing unseen relations and updating using existing graph structure. }\label{results_of_semi_unseen_random}
\vspace{-0.1cm}
\begin{tabular}{l|ccc|ccc|ccc|ccc}
\hline
\multicolumn{1}{c|}{\multirow{2}{*}{Methods}} & \multicolumn{3}{c|}{NELL-995.v1.v3}   & \multicolumn{3}{c|}{NELL-995.v2.v3}              & \multicolumn{3}{c|}{NELL-995.v4.v3}             & \multicolumn{3}{c}{FB15k-237.v1.v4}           \\
& AUC-PR & MRR & Hits@10  & AUC-PR & MRR & Hits@10 & AUC-PR  & MRR & Hits@10   & AUC-PR  & MRR             & Hits@10          \\\hline
TACT-base
& 73.98 & 43.59 & 72.48
& 85.88 & 65.63 & 83.47 
& 72.40 & 52.68 & 67.95
& \textbf{90.29} & \textbf{61.02} & \textbf{82.75}                   \\
RMPI-base
& 84.06 & \textbf{59.10}
& \textbf{82.12} & 89.98 & 67.98 & 82.30
& \textbf{88.20} & \textbf{70.33} & \textbf{81.20}   
& 88.76 & 56.81 & 79.71 \\
RMPI-NE
& \textbf{84.86} & 56.19 & 78.63
& \textbf{91.10} & \textbf{73.90} & \textbf{88.78} 
& 83.97 & 59.47 & 72.86   
& 88.99 & 57.77 & 80.38         \\
\hline
\end{tabular}
\end{subtable}
\newline
\vspace*{0.2cm}
\newline
\begin{subtable}{1\linewidth}
\centering
\caption{\textit{Schema Enhanced}: initializing unseen relations using vectors learned from ontological schemas (well support for NELL related datasets).}
\label{results_of_semi_unseen_schema}
\vspace{-0.1cm}
\begin{tabular}{l|ccc|ccc|ccc}
\hline
\multicolumn{1}{c|}{\multirow{2}{*}{Methods}}    & \multicolumn{3}{c|}{NELL-995.v1.v3}              & \multicolumn{3}{c|}{NELL-995.v2.v3}             & \multicolumn{3}{c}{NELL-995.v4.v3}           \\
& AUC-PR & MRR & Hits@10 & AUC-PR  & MRR & Hits@10   & AUC-PR  & MRR                  & Hits@10              \\\hline
TACT-base
& \textbf{93.20} & 68.30 & 89.28 
& 93.49 & 74.41 & 90.26  
& 93.53 & 74.89 & \textbf{90.25}      \\
RMPI-base
& 92.96 & 70.72 & 90.34 
& 94.31 & \textbf{80.37} & 92.05 
& \textbf{94.16} & \textbf{79.20} & 89.60          \\
RMPI-NE
& 92.50 & \textbf{72.24} & \textbf{91.08} 
& \textbf{94.99} & 80.03 & \textbf{92.94} 
& 93.44 & 72.69 & 88.53         \\
\hline

\end{tabular}
\end{subtable}%
\end{table*}

\subsection{Baselines and Model Variants}
\subsubsection{Baselines}
In addition to GraIL as a pioneering method on subgraph-based partially inductive KGC with unseen entities, we also make comparisons with two important state-of-the-art baselines under different settings.
The first one is TACT \cite{TACT}. It consists of an entity-based message passing module as GraIL and a relational correlation module which models the topological correlations between the target relations and their adjacent relations and applies a relational correlation network to enrich the representations of the target relations that are omitted by GraIL.
These two modules are integrated to compute the score of the target triple.
Also, the single relation correlation module alone can be used for triple scoring, which is reported as a base model named as TACT-base.
%In our paper, 
We report the results of both TACT and TACT-base for a comprehensive evaluation.
Note the original TACT generates negative samples in training and testing by replacing the relation of a triple, and predicts and ranks the relations for performance measurement. 
%them with the ground truth for relation prediction, 
We re-implement it with entity replacement and entity prediction as the other methods for fair comparisons.
The second baseline is CoMPILE \cite{CoMPILE}, a model that updates relation and entity embeddings simultaneously during message passing so as to strengthen the interaction between entities and relations.
The comparisons with traditional rule learning based methods are omitted as the poorer results than GraIL as reported in \cite{GraIL}.
%please refer to the results present in GraIL.

For fully inductive KGC, the above methods except for TACT-base assume that all the relations in the testing graph have been seen in the training graph and thus cannot be applied.
For TACT-base, the embedding of the unseen relation in a target triple can be 
%except for the TACT-base model in which the inductive embedding of an unseen target relation can be 
inferred by the relational correlation module via aggregating the embeddings of its adjacent relations.
% \todo{seen relations in the subgraph}.
Therefore, it is possible to make a comparison with TACT-base on the fully inductive KGC benchmarks.

Another important baseline we consider for fully inductive KGC is MaKEr \cite{MaKEr}, which also utilizes the local graph structure to deal with unseen entities and unseen relations. In MaKEr, the initial embeddings of unseen relations are represented by some pre-defined
%meta-relations (i.e., 
topological relationships between relations, while the initial embeddings of unseen entities are represented by the embeddings of their neighboring relations.
It adopts a meta learning framework which formulates a set of training tasks to mimic the testing graph and constructs a validation graph to tune the model parameters.
For convenience and fair comparison, we run our model on the benchmarks released in \cite{MaKEr}, i.e., FB-Ext and NELL-Ext --- two datasets respectively derived from FB15k-237 and NELL-995, with new validation sets which are generated by cleaning the 
original validation graph to ensure that all the entities and relations are seen in the training graph as we have set.
Statistically, the numbers of the validation triples filtered out in FB-Ext and NELL-Ext are $336$ and $137$, respectively.
See \cite{MaKEr} for more details of these two benchmarks.
It is worth noting that MaKEr falls into the fully inductive KGC setting of \textit{testing with semi unseen relations}, that is, its testing graphs include both seen and unseen relations. 
Following the original evaluation in MaKEr, the testing triples are divided into three sets:
%groups according to whether they are partially or fully inductive, i.e., 
\textit{(i)} \textit{u\_ent} where all the entities are unseen while all the relations are seen; \textit{(ii)} \textit{u\_rel} where all the entities are seen while all the relations are unseen; and \textit{(iii)} \textit{u\_both} where each triple to predict involves an unseen relation and at least one unseen entity.
We follow \cite{MaKEr} to report results on entity prediction with metrics of MRR and Hits@10.

\subsubsection{RMPI Variants}
To evaluate the effectiveness of two techniques --- the one-hop neighborhood aggregation in the disclosing subgraph for handling empty enclosing subgraphs (NE), and the target relation-aware neighborhood attention in message aggregation (TA), we tested several RMPI variants: RMPI-base without NE and TA, RMPI-NE with NE, RMPI-TA with TA, and RMPI-NE-TA with both NE and TA.

\begin{table*}[htbp]
\centering
\caption{\small Results of fully inductive KGC in the \textit{testing with fully unseen relations} setting. The best results are marked in bold.}
\label{results_of_fully_unseen}
\vspace{-0.1cm}
\begin{subtable}{1\linewidth}
\centering
\caption{\small \textit{Random Initialized}: randomly initializing unseen relations and updating using existing graph structure. }\label{results_of_fully_unseen_random}
\vspace{-0.1cm}
\begin{tabular}{l|ccc|ccc|ccc|ccc}
\hline
\multicolumn{1}{c|}{\multirow{2}{*}{Methods}} & \multicolumn{3}{c|}{NELL-995.v1.v3}   & \multicolumn{3}{c|}{NELL-995.v2.v3}              & \multicolumn{3}{c|}{NELL-995.v4.v3}             & \multicolumn{3}{c}{FB15k-237.v1.v4}           \\
& AUC-PR & MRR & Hits@10  & AUC-PR & MRR & Hits@10 & AUC-PR  & MRR & Hits@10   & AUC-PR  & MRR                  & Hits@10              \\\hline
TACT-base
& 73.47 & 34.84 & 52.64
& 77.78 & 40.87 & 68.75 
& 67.46 & 41.80 & 59.23       
& 55.52 & 15.16 & 15.82
\\
RMPI-base
& 83.14 & 52.78 & 79.09
& 86.18 & 52.48 & 70.00
& \textbf{82.09} & \textbf{55.14} & 65.21
& 58.77 & 13.74 & 15.57 \\
RMPI-NE
& \textbf{84.40} & \textbf{55.39} & \textbf{82.57}
& \textbf{87.93} & \textbf{61.01} & \textbf{81.51}
& 79.38 & 50.62 & \textbf{67.10}
& \textbf{60.23} & \textbf{21.36} & \textbf{25.82}
\\
\hline
\end{tabular}
\end{subtable}
\newline
\vspace*{0.2cm}
\newline
\begin{subtable}{1\linewidth}
\centering
\caption{\small \textit{Schema Enhanced}: initializing unseen relations using vectors learned from ontological schemas (well support for NELL related datasets).}
\label{results_of_fully_unseen_schema}
\vspace{-0.1cm}
\begin{tabular}{l|ccc|ccc|ccc}
\hline
\multicolumn{1}{c|}{\multirow{2}{*}{\revise{Methods}}}    & \multicolumn{3}{c|}{NELL-995.v1.v3}              & \multicolumn{3}{c|}{NELL-995.v2.v3}             & \multicolumn{3}{c}{NELL-995.v4.v3}           \\
& AUC-PR & MRR & Hits@10 & AUC-PR  & MRR & Hits@10   & AUC-PR  & MRR                  & Hits@10              \\\hline
TACT-base
& 92.66 & 68.55 & 89.09
& 91.87 & 62.19 & 86.70
& 90.80 & 62.54 &80.72
\\
RMPI-base
& 92.56 & 70.00 & 90.23
& 93.29 & 73.36 & 90.20
&
\textbf{91.26}   
&\textbf{69.47} & 81.33        \\
RMPI-NE
& \textbf{92.95} & \textbf{71.82} & \textbf{91.18}  
& \textbf{94.19} & \textbf{74.80} & \textbf{92.09} 
& 90.09 & 64.25 & \textbf{81.44} \\
\hline
\end{tabular}
\end{subtable}%
\end{table*}

\begin{table*}
\caption{\small Comparison results with MaKEr \cite{MaKEr} on its developed benchmarks (unseen relations are randomly initialized in our models).}
\label{results_with_maker_random}
\vspace{-0.1cm}
  \centering
\begin{tabular}{l|cc|cc|cc|cc|cc|cc}
\hline
\multicolumn{1}{c|}{\multirow{3}{*}{Methods}} & \multicolumn{6}{c|}{FB-Ext}             & \multicolumn{6}{c}{NELL-Ext}\\
&    \multicolumn{2}{c|}{u\_ent} & \multicolumn{2}{c|}{u\_rel} & \multicolumn{2}{c|}{u\_both} &
\multicolumn{2}{c|}{u\_ent} & \multicolumn{2}{c|}{u\_rel} & \multicolumn{2}{c}{u\_both}\\

& MRR &  Hits@10 &  MRR &  Hits@10 &  MRR & Hits@10
&  MRR &  Hits@10 &  MRR &  Hits@10 &  MRR &  Hits@10
\\
\hline
% Asmp-KGE        
% & 63.91 & 82.22 & 34.61 & 36.50 
% & 13.29 & 25.24 & 68.64 & 78.35
% & 14.32 & 22.50 & 9.94  & 17.48  \\
MaKEr
& \textbf{74.64} & \textbf{95.28} & 32.00 & 50.00 
& 27.26 & 52.09 & \textbf{77.09} & 94.64 
& 31.53 & 60.00 & 41.39 & 62.35 \\
RMPI-base
& 50.48 & 82.55
& \textbf{51.61} & \textbf{69.50}
& 30.54 & 47.73
& 76.36 & 94.85
& 46.09 & 82.50
& 34.82 & 63.48
\\ 
RMPI-NE 
& 53.51 & 82.85
& 51.47 & 69.00
& \textbf{34.38} & \textbf{56.77}  
& 75.46 & \textbf{96.05} 
& \textbf{63.99} & \textbf{89.17}   
& \textbf{42.91} & \textbf{71.48}  
\\ 
\hline
\end{tabular}
\end{table*}

\begin{table}[htbp]
\centering
\caption{\small Comparison results with MaKEr \cite{MaKEr} on NELL-Ext when our models are enhanced by ontological schema.}\label{results_with_maker_schema}
\vspace{-0.1cm}
\resizebox{\linewidth}{!}{
\begin{tabular}{l|cc|cc|cc}
\hline
\multicolumn{1}{c|}{\multirow{2}{*}{Methods}}
&
\multicolumn{2}{c|}{u\_ent} & \multicolumn{2}{c|}{u\_rel} & \multicolumn{2}{c}{u\_both}\\
&  MRR &  Hits@10 &  MRR &  Hits@10 &  MRR &  Hits@10
\\
\hline
MaKEr
& \textbf{77.09} & 94.64 
& 31.53 & 60.00 & 41.39 & 62.35
\\
RMPI-base &
76.52 & 94.92  
& 66.23 & 95.83   
& 43.04 & 75.48
\\
RMPI-NE &
75.90 & \textbf{95.19} 
& \textbf{75.16} & \textbf{99.17} 
& \textbf{46.04} & \textbf{75.83}
\\ 
\hline
\end{tabular}}
\end{table}
\subsection{Main Results of Fully Inductive KGC}
We have two settings to initially represent an unseen relation in the testing graph: \textit{(i)} random initialization, and \textit{(ii)} projection from the relation's semantic vector learned from the ontological schema.
These two settings are denoted as \textit{Random Initialized} and \textit{Schema Enhanced}, respectively.
Both of them are tested with TACT-base, RMPI-base and RMPI-NE, while RMPI-TA and RMPI-NE-TA cannot be applicable here.
% \todo{Since random initialization may cause a negative impact on the target relation-aware attention, we only report the results of RMPI-base and RMPI-NE here.}

\subsubsection{Test with semi unseen relations}
%\textup{We first compare with TACT-base in the} testing with semi unseen relations \textup{setting on our constructed fully inductive benchmarks, the results are reported in \textup{Table~\ref{results_of_semi_unseen}}}
The results of fully inductive KGC under the test setting of semi unseen relations are shown in Table~\ref{results_of_semi_unseen}.
From Table \ref{results_of_semi_unseen_random}, we can observe that our models always outperform TACT-base under the setting of \textit{Random Initialized}.
This shows the effectiveness of our proposed relational message passing network which passes messages between relations with multiple aggregation layers, through which the inductive embeddings of unseen relations can be well learned by aggregating their neighboring seen relations.
In contrast, the relation correlation module in TACT-base actually only performs one-layer neighborhood aggregation for the central target relation, while for unseen relations that exist in the neighborhood, it cannot provide meaningful representations.
The only exception raised in FB15k-237.v1.v4 may be because that \textit{(i)} the proportion of unseen relations in its testing graph is relatively low ($174$ seen relations vs. $26$ unseen relations), where the negative impact of randomly initialized unseen relations is relatively limited;
and \textit{(ii)} the one-hop neighboring relations are informative enough for predicting the triple.
%in FB15k-237.v1.v4.

When the representations of the unseen relations are augmented by the ontological schema, the performance of both our models and TACT-base is greatly improved, as shown in Table \ref{results_of_semi_unseen_schema}.
This validates our insight of enriching the neighborhood connections between seen and unseen relations by ontological schemas.
Meanwhile, RMPI-base and RMPI-NE both achieve better or comparable performance against TACT-base, which is consistent with the results under \textit{Random Initialized}.

\subsubsection{Test with fully unseen relations}
The results of fully inductive KGC under the test setting of fully unseen relations are shown in Table \ref{results_of_fully_unseen}.
%\textup{We further make a comparison with TACT-base in a more extreme setting of} testing with fully unseen relations \textup{where all the relations in the test graph are unseen, the results are reported in Table \ref{results_of_fully_unseen}}}
It can be seen that the performance dramatically drops when performing random initialization as there are no seen relations around to provide meaningful information for updating the representations of unseen relations.
The only clue for reasoning may lie in the high-level graph patterns, e.g., the co-occurence patterns of relations, while our proposed models are better at capturing such patterns than TACT-base, which can be verified by the less performance degradation, especially on NELL-995.v1.v3 and NELL-995.v2.v3.

\subsubsection{Comparison with MaKEr}
%\textup{The comparisons with MaKEr on its developed benchmarks are presented, with the result of \textit{Random Initialized} reported in Table \ref{results_with_maker_random} and the result of \textit{Schema Enhanced} reported in Table \ref{results_with_maker_schema}, we follow MaKEr to report results on entity prediction with metrics of MRR and Hits@10}}
The results of \textit{Random Initialized} are reported in Table \ref{results_with_maker_random} while the results of \textit{Schema Enhanced} are reported in Table \ref{results_with_maker_schema}.
On the testing sets of \textit{u\_rel} and \textit{u\_both}, our models always achieve better results than MaKEr, even though the unseen relations are randomly initialized, and the performance is further improved when augmented by the ontological schema.
While on the testing set of \textit{u\_ent}, 
%a group of testing triples involving only unseen entities and seen relations, 
our models gain slightly superior Hits@10 and comparable MRR on NELL-Ext, but gain largely inferior MRR and Hits@10 on FB-Ext.
This may be attributed to \textit{(i)} the high coverage of seen relations in MaKEr's testing graphs, that is, there are $6103$ (resp. $2160$) triples in the testing graph of FB-Ext (resp. NELL-Ext) used for providing the local graph structures, $5713$ (resp. $1923$) out of which contain seen relations, thus the testing on \textit{u\_ent} is actually a setting close to the partially inductive KGC; and \textit{(ii)} the nature of relations in FB15k-237, i.e., the directed neighbors are informative enough to predict the triple.
Overall, our proposed message network accomplishes promising results in the fully inductive case compared with the mete learning based MaKEr.

\begin{table*}
\centering
\caption{\small Results of partially inductive KGC with only unseen entities. The best are in bold and the second best are underlined.}
\label{results_of_partially}
\vspace{-0.1cm}
\begin{subtable}{1\linewidth}
\centering
  \caption{\small Entity Prediction with Hits@10.}
  \label{results_partially_LP}
  \centering
  \vspace{-0.1cm}
  \begin{tabular}{c|cccc|cccc|cccc}
\hline
\multirow{2}{*}{Methods} & \multicolumn{4}{c|}{WN18RR} & \multicolumn{4}{c|}{FB15k-237} & \multicolumn{4}{c}{NELL-995} \\
& v1    & v2   & v3   & v4   
& v1    & v2    & v3    & v4    
& v1    & v2    & v3    & v4   \\
\hline
% Neural-LP
% & 74.37 & 68.93 & 46.18 & 67.13 
% & 52.92 & 58.94 & 52.90 & 55.88
% & 40.78 & 78.73 & 82.71 & 80.58 \\
% DRUM
% & 74.37 & 68.93 & 46.18 & 67.13 & 52.92 & 58.73
% & 52.90 & 55.88 & 19.42 & 78.55 & 82.71 & 80.58
%  \\
% RuleN                    
% & 80.85 & 78.23 & 53.39 & 71.59 
% & 49.76 & 77.82 & \textbf{87.69} & 85.60 
% & 53.50 & 81.75 & 77.26 & 61.35  \\
% \midrule
GraIL                   
& 82.45 & 78.68 & 58.43 & 73.41 
& 64.15 & 81.80 & 82.83 & \underline{89.29}
& \underline{59.50} & 93.25 & 91.41 & 73.19   \\
\hline
TACT-base           
& 82.45 & 78.68 & 58.84 & 73.34 
& 64.88 & 82.43 & \textbf{86.30} & \textbf{89.71} 
& 56.50 & 92.44 & 94.38 & 63.27      \\
% TACT-base + schema   
% &  &  &  & 
% & 63.66  &  &  & 
% &  & 92.44 &  & 84.88  \\
TACT
& 82.45 & 78.68 & 58.60 & 73.41 
& 62.20 & 80.02 & 84.16 & 88.41   
& 51.50 & 91.49 & 92.46 & 72.98    \\
\hline
CoMPILE                  
& 83.60 & 79.82 & 60.69 & 75.49   
& 67.66 & \underline{82.98} & 84.67 & 87.44    
& 58.38 & \underline{93.87} & 92.77 & 75.19     
\\
\hline
RMPI-base
& 82.45 & 78.68 & 58.68  & 73.41
& 65.37 & 81.80 & 81.10  & 87.25
& \underline{59.50} & 92.23 & 93.57 & \textbf{87.62} 
\\ 
% RMPI-NE
% & \textbf{89.36}  & \textbf{84.47}  &  \textbf{72.31} 
% & \textbf{80.83} 

% & \textbf{69.76}  & \underline{82.95}
% & 82.14 & 84.76 

% & \textbf{63.00} & 92.12 & \underline{93.02} & \textbf{83.52} \\
RMPI-NE
& \textbf{89.63} & \textbf{83.22} & \underline{70.33}  & \underline{79.81}  
& \underline{70.00} & 82.85 & 83.18  & 86.52 
& \textbf{60.50} & \textbf{94.01} & 91.78  & \underline{84.27} 
% \\
% RMP-base + schema   
% &  &  &  & 
% & 72.44  &  &  & 
% &  & 90.44 &  & 82.90
\\ 
RMPI-TA
& 82.45 & 78.68 & 58.84 & 73.41
& 66.10 & 82.53 & 84.74 & 87.78
& 53.00  & 93.17 & \textbf{95.36} & 45.62 
\\ 
RMPI-NE-TA
& \underline{87.77} & \underline{82.43}  & \textbf{73.14} & \textbf{81.42}  
& \textbf{71.71} & \textbf{83.37}  & \underline{86.01}  & 88.69
& \textbf{60.50} & 93.49 & \underline{95.30} & 66.42
\\ 
\hline
\end{tabular}
\end{subtable}
\newline
\vspace*{0.2cm}
\newline
\begin{subtable}{1\linewidth}
\centering
  \caption{\small Triple Classification with AUC-PR.}
  \label{results_partially_TC}
  \vspace{-0.1cm}
  \centering
  \begin{tabular}{c|cccc|cccc|cccc}
\hline
\multirow{2}{*}{Methods} & \multicolumn{4}{c|}{WN18RR} & \multicolumn{4}{c|}{FB15k-237} & \multicolumn{4}{c}{NELL-995} \\
& v1    & v2   & v3   & v4   
& v1    & v2    & v3    & v4    
& v1    & v2    & v3    & v4   \\
\hline
% Neural-LP    
% & 86.02 & 83.78 & 62.90 & 82.06 
% & 69.64 & 76.55 & 73.95 & 75.74 
% & 64.66 & 83.61 & 87.58 & 85.69 \\
% DRUM
% & 86.02 & 84.05 & 63.20 & 82.06 
% & 69.71 & 76.44 & 74.03 & 76.20 
% & 59.86 & 83.99 & 87.71 & 85.94     \\
% RuleN                    
% & 90.26 & 89.01 & 76.46 & 85.75
% & 75.24 & 88.70 & 91.24 & 91.79
% & \underline{84.99} & 88.40 & 87.20 & 80.52   \\
% \midrule
GraIL                   
& 94.32 & 94.18 & 85.80 & 92.72 
& 84.69 & 90.57 & 91.68 & 94.46 
& \textbf{86.05} & 92.62 & 93.34 & 87.50     \\
\hline
TACT-base    
& \underline{96.89} & 97.40 & 88.43 & 97.08  
& \textbf{86.24} & \textbf{93.12} & \textbf{94.83} & \textbf{95.09}  
& 80.75 & \underline{94.84} & 94.69 & 76.42 \\
% TACT-base + schema   
% &  &  &  & 
% & 84.94  &  &  & 
% &  & 95.04 &  & 92.84  \\
TACT
& 96.27 & \underline{97.69} & 88.33 & \underline{97.26} 
& 85.03 & 91.72 & \underline{93.14} & 93.85
& 77.54 & 93.30 & 92.53 & 85.25      
\\
\hline
CoMPILE                  
& \textbf{98.23} & \textbf{99.56} & \textbf{93.60} & \textbf{99.80}   
& 85.50 & 91.68 & 93.12 & \underline{94.90}   
& 80.16 & \textbf{95.88} & \underline{96.08} & 85.48      \\
\hline
RMPI-base
& 95.00 & 95.96 & 88.53 & 95.78  
& 85.25 & 92.19 & 92.09 & 92.80
& \underline{81.12} & 93.46 & 95.35  & \textbf{91.77} 
\\ 
RMPI-NE
& 95.09 & 95.43 & \underline{88.58} & 94.82
& 85.22 & 92.08 & 91.77 & 92.27
& 81.07 & 93.64 & 94.99 & \underline{88.82}
\\ 
RMPI-TA
& 95.54 & 97.52 & 88.45 & 96.95  
& \underline{86.18} & 92.87 & 92.82 & 93.81 
& 78.24 & 94.58  & \textbf{96.77} & 69.11 
\\ 
RMPI-NE-TA
 
& 95.05 & 95.48 & 88.35 & 94.87 
& 85.90 & \underline{92.96}  & 92.72  & 93.33 
& 77.89 & 94.31 & 95.89 & 72.34 
\\ 
% RMP-base + schema   
% &  &  &  & 
% & 85.02  &  &  & 
% &  & 92.96 &  & 84.19  \\
\hline
\end{tabular}
\end{subtable}%
\end{table*}

\subsection{Main Results of Partially Inductive KGC}\label{}
Our method RMPI is specifically developed for fully inductive KGC, but is also compatiable to partially inductive KGC.
Thus we also evaluate the variants of RMPI for partially inductive KGC by comparing them with the state-of-the-art subgraph reasoning-based methods including GraIL, TACT, TACT-base and CoMPILE.
%partially inductive reasoning methods.
The results of entity prediction are shown in Table \ref{results_partially_LP}.
We can see that our models always outperform the baselines.
%in ranking ground-truth entities higher than other candidates.
Especially, our models achieve a large margin of improvements on the series of datasets from WN18RR and NELL-995.v4.
%are significant, where our models outperform the SOTA by a large margin.
The results of triple classification are shown in Table \ref{results_partially_TC}.
Although our models are not dominated, they are still the second best or comparable to the second best.
These results indicate the superiority of our relational message passing network against these baselines which mainly rely on the entities for message passing.
\revise{Moreover, regarding the less space for improvement in triple classification task (i.e., the AUC-PR values are more than 90\% and close to 100\% in most situations), we prefer to claim our superiority on the more challenging entity prediction task.}
%and highlight the necessity of taking relations as an important clue for inductive reasoning.

\subsection{Ablation Studies}

\subsubsection{Different Components}
We investigate the contributions of the one-hop neighborhood aggregation in disclosing subgraphs (NE) and the target relation-aware neighborhood attention in message passing (TA) by comparing the variants of RMPI.
% \todo{We mainly take the results in the partially inductive KGC for a detailed analysis.}
First, 
for NE, we find that RMPI-NE improves RMPI-base in most situations in the fully inductive KGC cases, especially in \textit{testing with fully unseen relations} with \textit{Random Initialized}, illustrating that the additional inputs from the disclosing graph provide more graph patterns for prediction.
%in such an extreme setting.
The improvements vary from case to case, this could be attributed
%While the overall inconsistent improvement may be attributed 
to the datasets and their triples for prediction. The information from the enclosing subgraph sometimes is enough to make a reasonable prediction.
In the partially inductive case,  RMPI-NE and RMPI-NE-TA outperform RMPI-base and RMPI-TA, respectively, on either entity prediction or triple classification in most situations, as shown in Table \ref{results_partially_LP} and Table \ref{results_partially_TC}.
In particular, for the entity prediction on WN18RR derived benchmarks, where many negative triples and quite a few positive triples have empty enclosing subgraphs, NE plays a significant positive role, e.g., RMPI-NE-TA (RMPI-NE resp.) achieves an averaged $7.85\%$ ($7.40\%$ resp.) improvement over RMPI-TA (RMPI-base resp.).
To sum up, with the technique of NE, the disclosing graph is able to provide discriminative features when the enclosing subgraph are absent, and complementary semantics when the enclosing subgraph exists.

% \todo{Both RMPI-base and RMPI-NE advance our framework to achieve promising performance in fully inductive KGC, the inconsistent effectiveness of the additional inputs from the disclosing subgraphs may be due to the nature of datasets, that is, in some cases, the }

Second, we verify the effectiveness of TA by comparing  RMPI-base against RMPI-TA, and RMPI-NE against RMPI-NE-TA. 
%the performance of variants that are with or without this component, i.e., RMPI-base vs. RMPI-TA, and RMPI-NE vs. RMPI-NE-TA.
From Table \ref{results_partially_LP} and Table \ref{results_partially_TC}, we can observe that the variants accompanied by TA perform better in most situations.
For example, on the benchmarks derived from FB15k-237, the average improvement of RMPI-TA over RMPI-base on entity prediction is $1.41\%$, and that of RMPI-NE-TA over RMPI-NE is $1.81\%$. On triple classification, the corresponding improvements are $0.84\%$ and $0.89\%$, respectively.
These results suggest that it is necessary to consider the importance of different neighboring relations w.r.t. the target relation, especially for those relations that are more than one hop away.
%from the target relation.
However, this mechanism may not work well for some datasets 
%regarding the poor performance on 
such as NELL-995.v4.
It may be because some distant relations that are important for reasoning are filtered out due to their less relatedness to the target relation.
In the future, we will study more robust mechanisms for TA.
%to determine which relations in the neighborhood are more important in a more suitable way.

\subsubsection{Different Fusion Functions}
To integrate the information from the enclosing and disclosing subgraphs, we design the summation-based fusion function \eqref{eq:sum_combine} and the concatenation-based fusion function \eqref{eq:conc_combine}.
Both functions can advance our model for better performance (see the analysis on RMPI-NE).
%two combination functions, i.e., summation-based and concatenation-based, as introduced in Section \ref{method_NE}.
Table \ref{ablation_fusion}  presents the results of these two functions in different cases for comparison.
%for a comprehensive comparison, as Table \ref{ablation_fusion} shows.
We can find that the better function varies from dataset to dataset, from setting to setting and from task to task.
In the future, we will investigate more robust fusion functions that can generalise to different situations.
%different datasets as well as different settings.
% \todo{Still, both of them positively contribute the fusion of  so that more discriminative features are provided for triple scoring.}

\subsubsection{Ontological Schemas}
In fully inductive KGC, the extra relation semantics from ontological schemas consistently improve the performance of RMPI as we have presented.
%over that is solely based on the existing structure, 
%as the results we introduced earlier.
While in partially inductive KGC, we also investigate the impact of using ontological schemas.
%modeling richer connections between seen relations.
The results of our models and the baseline TACT-base, with and without schemas, on two benchmarks derived from NELL-995 are shown in Table \ref{results_partially_onto}.
We can observe that the performance is augmented in most situations, especially for TACT-base on NELL-995.v4.
This large improvement on NELL-995.v4 may be due to the compact relatedness of the relations.
TACT-base only models the part of relation correlations, while the ontological schema provides relatively important complementary relation semantics for it.
%the nature of relations in NELL-995.v4, i.e., they are strongly related to each other. 
%When TACT-base only models the correlations between the target relations and their adjacent relations, the extra relations connections provided by ontological schemas significantly enlarge the reasoning space so that greatly improving its reasoning ability.
All of these results illustrate the great signiﬁcance of utilizing external relation semantics from the ontological schema.
%utilizing the additional relation semantics to augment the subgraph reasoning with relations. 

% \revise{\subsubsection{Computation Efficiency} 
% For subgraphs of different sizes, we also count the processing time
% %on processing them for deep analysis, especially w.r.t. 
% of different models. Specifically, we use the number of edges in the entity-view subgraph (i.e., the number of nodes in the transformed relation-view subgraph) to describe the graph size, and run RMPI-base, RMPI-TA and RMPI-NE (with summation-based fusion function) on CPU to test the triples in the partially inductive setting with subgraph sizes 
% %around such as 
% of around $100$, $1000$, $5000$ and $20000$.
% The averaged inference time (seconds) of RMPI-base on graphs of these four scales are 0.031, 0.053, 0.132 and 7.131, respectively, that of RMPI-TA are 0.036, 0.058, 0.202, and 21.311, respectively, and that of RMPI-NE are 0.053, 0.059, 0.159 and 8.539, respectively.
% %According to these statistics, 
% We can find that \textit{i}) the running time increases as the graph size increases and the complexity of models increases; \textit{ii}) the time gap between RMPI-base and RMPI-TA is greatly enlarged as the graph size increases, especially when the size reaches 20000. In the future, we will test the inference time using GPU.}

\begin{table}
\centering
\caption{\small Results of RMPI-NE using the summation-based fusion function (SUM) function and the concatenation-based fusion function (CONC).
$^{\dag}$: the results under \textit{testing with semi unseen relations}.
}\label{ablation_fusion}
\vspace{-0.1cm}
\begin{subtable}{1\linewidth}
\centering
\caption{\small Partially inductive KGC}
\vspace{-0.1cm}
\resizebox{\linewidth}{!}{
\begin{tabular}{l|cc|cc|cc}
\hline
\multicolumn{1}{c|}{\multirow{2}{*}{Function}} & \multicolumn{2}{c|}{NELL-995.v2}   & \multicolumn{2}{c|}{NELL-995.v4}& \multicolumn{2}{c}{FB15k-237.v1}        \\
&  AUC-PR &  Hits@10  &  AUC-PR &  Hits@10 
&   AUC-PR &  Hits@10       \\\hline
SUM
& \textbf{93.64} & 91.81 
& 88.24 & \textbf{84.27}  
& 84.98 & \textbf{70.00}  
\\
CONC
& 93.52 & \textbf{94.01}
& \textbf{88.82} & 79.07
& \textbf{85.22} & 66.34 
\\
\hline
\end{tabular}}
\end{subtable}
\newline
\vspace*{0.1cm}
\newline
\begin{subtable}{1\linewidth}
\centering
\caption{\small Fully inductive KGC$^{\dag}$ (\textit{Random Initialized})}
\vspace{-0.1cm}
\resizebox{\linewidth}{!}{
\begin{tabular}{l|cc|cc|cc}
\hline
\multicolumn{1}{c|}{\multirow{2}{*}{Function}} & \multicolumn{2}{c|}{NELL-995.v2.v3}   & \multicolumn{2}{c|}{NELL-995.v4.v3}              & \multicolumn{2}{c}{FB15k-237.v1.v4}            \\
&  AUC-PR &  Hits@10  &  AUC-PR &  Hits@10 
&   AUC-PR &  Hits@10      \\\hline
SUM
& 90.21 & 83.26 
& \textbf{83.97}  & 72.86  
& \textbf{88.99} & \textbf{80.38}  
\\
CONC
& \textbf{91.10} & \textbf{88.78}
& 82.23 & \textbf{73.56}
& 87.26 & 75.43
\\
\hline
\end{tabular}}
\end{subtable}%
\newline
\vspace*{0.1cm}
\newline
\begin{subtable}{1\linewidth}
\centering
\caption{\small Fully inductive KGC$^{\dag}$ (\textit{Schema Enhanced})}
\vspace{-0.1cm}
\scriptsize
\begin{tabular}{l|cc|cc}
\hline
\multicolumn{1}{c|}{\multirow{2}{*}{Function}} & \multicolumn{2}{c|}{NELL-995.v2.v3}
&\multicolumn{2}{c}{NELL-995.v4.v3}           \\
&  AUC-PR &  Hits@10
&   AUC-PR &  Hits@10       \\\hline
SUM
& \textbf{94.99} & \textbf{92.94} 
& 86.24 & 72.24 
\\
CONC
& 94.56 & 92.67 
& \textbf{93.44} & \textbf{88.53}
\\
\hline
\end{tabular}
\end{subtable}%
\end{table}

\begin{table}
\centering
\caption{\small Partially inductive KGC with (w) and without (w/o) ontological schemas. (S) and (C) represent the summation-based and concatenation-based fusion functions, respectively. The results improved by the ontological schemas are underlined.}
\label{results_partially_onto}
% \vspace{-0.1cm}
% \resizebox{\linewidth}{!}{
\begin{tabular}{p{0.3cm}<{\centering}|p{1.35cm}|p{0.9cm}<{\centering}p{0.4cm}<{\centering}p{0.7cm}<{\centering}|p{0.9cm}<{\centering}p{0.4cm}<{\centering}p{0.7cm}<{\centering}}
% \begin{tabular}{c|l|ccc|ccc}
\hline
\multicolumn{1}{c|}{\multirow{2}{*}{}}    &
\multicolumn{1}{c|}{\multirow{2}{*}{Methods}}                 & \multicolumn{3}{c|}{NELL-995.v2}             & \multicolumn{3}{c}{NELL-995.v4}           \\
&  &  \scriptsize AUC-PR  &   \scriptsize MRR &    \scriptsize Hits@10   &    \scriptsize AUC-PR  &   \scriptsize  MRR          &  \scriptsize  Hits@10        \\\hline
\multirow{4}{*}{w/o} &\scriptsize TACT-base
% & 80.75 & 49.76 & 56.50 
& 94.84 & 75.06 & 92.44 
& 76.42 & 56.67 & 63.27 \\
& \scriptsize RMPI-base
% & 81.12 & 53.43 & 59.50 
& 93.46 & 76.65 & 92.23 
& 91.77 & 74.68 & 87.62                  \\
& \scriptsize RMPI-NE(S)
% &  &  & 
& 93.64 & 75.48 & 91.81 
& 88.24 & 64.36 & 84.27            \\
&  \scriptsize RMPI-NE(C)
% &  &  & 
& 93.52 & 76.29 & 94.01 
& 88.82 & 66.93 & 79.07        \\
\hline
\multirow{4}{*}{w} & \scriptsize TACT-base
% & 79.24 & 52.39 & 60.00 
& 94.40 & 74.09 & 90.34 
& \underline{91.13} & \underline{69.45} & \underline{85.77} \\
& \scriptsize RMPI-base
% & 80.47 & 52.59 & 62.50 
& \underline{94.04} & 75.54 & \underline{93.28} 
& \underline{92.17} & \underline{75.90} & 87.00          \\
& \scriptsize RMPI-NE(S)
% & 79.64 &  & 
& \underline{94.43} & \underline{76.86} & \underline{93.59} 
& \underline{88.78} & 63.95 & 82.97        \\
& \scriptsize RMPI-NE(C)
% & 79.64 &  & 
& 91.90 & 72.47 & 93.28
& \underline{91.42} & \underline{67.73} & \underline{85.13}                       \\
\hline
\end{tabular}
\end{table}

\subsection{Case Studies}
\begin{figure*}
  \centering
  \includegraphics[width=0.9\linewidth]{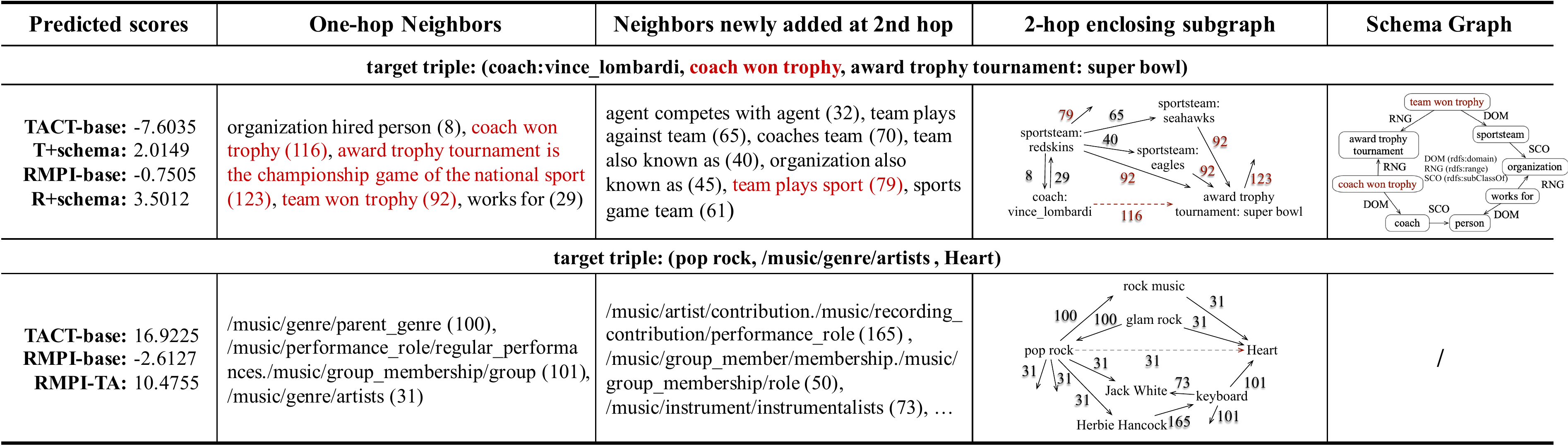}
\caption{\small Two positive target triples from NELL-995.v4.v3 and FB15k-237.v1.v4, respectively. Unseen relations are colored in red. The number behind each relation denotes its index (ID) in the dataset, and is used to label the edge in the enclosing graph here.
%and is used to label the edges in the enclosing graph. 
``T+schema'' and ``R+schema'' denote ontological schema enhanced TACT-base and RMPI-base, respectively.}
\label{fig:case_study}
\end{figure*}

In Fig. \ref{fig:case_study}, we present two positive target triples to be predicted, as well as their enclosing subgraphs, the relations in their neighborhoods, and their scores predicted by different models.
%from NELL-995 and FB15k-237 derived benchmarks, and show their predicted scores by different models, 
%extracted enclosing subgraphs and relations in the neighborhood, as we as the ontological schema segments for providing extra relation semantics in Fig. \ref{fig:case_study}.
%
%The first triple includes an unseen relation \textit{coach won trophy}. RMPI-base predicts higher score than TACT-base, while the schema enhance RMPI-based predicts the highest score.
%We can find there are both seen and unseen relations in the relation's neighbourhood, which verifies the capability of RMPI for fully inductive KGC. For example, \textit{team also known as} and \textit{team against team} could be utilized to imply that \textit{team won trophy} is a team related relation and further infer that a team and its member (could implied by \textit{works for}) tend to own the same thing.
%The ontology schema also connects the seen and unseen relations by the domain and range relationships to concepts.
The first example includes an unseen relation \textit{coach won trophy}.
RMPI-base predicts higher score than TACT-base, and the schema enhanced RMPI-base predicts the highest score.
As we can see, there are both seen and unseen relations in the neighborhood, while our models are more robust to perform reasoning.
More specifically, the one-hop neighboring relation \textit{team won trophy} that could be used to infer the target relation together with \textit{works for} is also an unseen relation whose embedding is not available, while the 2-hop seen relations such as \textit{team also known as} and \textit{team against team} that can be utilized by our models could imply that \textit{team won trophy} is a \textit{team} related relation and may further infer that a team and its one member (implied by \textit{works for}) tend to own the same thing.
When referring to the ontological schema where the seen and unseen relations are connected to some shared concepts via e.g. domain and range relationships, the real meanings of the unseen relations can be figured out, and more spaces are exposed for reasoning, thus resulting in improved prediction scores of TACT-base and RMPI-base.
From the second example, we can see that the one-hop neighboring relations \textit{/music/genre/parent\_genre} and \textit{/music/genre/artists} are informative enough to make a decision, that is, an artist falling in the child music genre also is labeled by the parent genre, while the 2-hop neighbors are noisy to collaborate.
As a result, TACT-base performs better on predicting it than RMPI-base.
However, the target relation-aware attention mechanism applied in the RMPI-TA improves the score of the second triple by highlighting the relations in the remote neighborhood and filtering those that are noisy for prediction.

\section{Related Work}\label{related_work}
\subsection{Transductive Knowledge Graph Completion}
Existing transductive KG completion methods usually assign a unique vector to each entity and relation and train it using the observed triple facts.
During evaluation, the embeddings of entities and relations in a testing triple are then looked up to score its plausibility.
% In general, the valid triples have higher scores than those invalid ones.
According to how a triple is scored, these methods can be grouped into three categories: \textit{translation}-based ones such as TransE \cite{transe}, TransH \cite{transh} and RotatE \cite{rotate}, \textit{semantic matching}-based ones such as DistMult \cite{distmult} and ComplEx \cite{complex}, and \textit{neural network}-based ones such as ConvE \cite{ConvE}.
% and NTN \cite{NTN}. 
Recently, inspired by the great success of graph neural networks (GNNs) in processing graph structured data,
%graph data modeling, 
some works employ relation-aware GNNs as powerful encoders to encode the structural information contained in KGs.
Typical practices include RGCN \cite{RGCN}, CompGCN \cite{CompGCN}, etc.
All of these models have shown their capability to capture complex semantic patterns in KGs and achieve state-of-the-art completion results in the transductive setting.
%However, they all fail to complete triples in the inductive setting, where the embeddings of newly emerging entities and relations are out of the embedding table. 
However, they fail to complete triples involving unseen entities and/or relations that have never appeared in the triples used for training.

% KG completion (KGC) usually refers to the task of predicting plausible triples (relational facts) that are not asserted in a KG.
% The methods often score triples according to the embeddings (vector representations) of the entities and relations, which are learned from the asserted triples.  
% Transductive KGC refers to a simple and idea setting where the new triples to predict only involve entities and relations that have already appeared in the triples used for training. 

% Such emerging entities and relations are quite common in real-word KGs which are usually involving or incrementally populated. 

% our methods: only rely on relation information for propagation and in inductive setting.

\subsection{Inducitve Knowledge Graph Completion}

\subsubsection{Partially Inductive}
The partial inductive KGC settings address either the unseen entities or unseen relations.
Several lines of inductive KGC methods have been proposed to learn embeddings for unseen entities following different settings \cite{chen2021low}.

One line of them use entity's external resources such as text  and images as additional inputs \cite{chen2021knowledge,xie2016representation,shah2019open,IKRL}.
For example, to incorporate with the textual information, early works first learn a continuous text embedding for each entity by e.g. averaging pre-trained word embeddings \cite{shah2019open} or applying deep convolutional neural models \cite{xie2016representation} for tokens in its name or description, and then map the text embeddings to compute the triple score together with the geometric embeddings learned by traditional transductive KGC methods.
The embeddings of unseen entities can thus be obtained via applying the learned mapping.
Instead of learning two kinds of representations, some recent works directly use the text resources to predict triples.
For example, KG-BERT \cite{kgbert} and StAR \cite{wang2021structure}, which build upon pre-trained language models such as BERT, model the KGC as a downstream sequence classification task, mainly utilizing the literal labels or descriptions of the entities.

Another line of works consider a few auxiliary triples of an emerging entity.
These triples link emerging entities to seen entities that have known to the KG and serve as the context to compute their embeddings.
For example, Hamaguchi et al. \cite{hamaguchi2017knowledge} apply a powerful GNN to generate the embedding of a new entity by aggregating all its neighboring known entities.
The subsequent works \cite{LAN,bhowmik2020explainable,baek2020learning} propose different neighborhood aggregation functions for optimizing this process.

Different from the above works which rely on entity's external resources or auxiliary triples, 
some efforts have been made to process the unseen entities solely based on the existing graph structure by inducing probabilistic logical rules from the KG, which capture entity-independent relational semantics and are inherently inductive \cite{meilicke2018fine,NeuralLP,DRUM,RuleN}.
In particular, Teru et al. \cite{GraIL} propose GraIL which aims to implicitly induce the logical rules by reasoning over the the enclosing subgraph surrounding a candidate triple in an entity-independent manner.
The follow-up works improve this process by introducing the topological correlations between relations \cite{TACT} or optimizing the message passing in a node-edge interaction manner \cite{CoMPILE}.
While in this paper, we focus on the information of relations and conduct relational message passing to predict triples. 

However, the methods introduced above mainly aim at unseen entities and require all the testing relations have been seen at training stage.
There are also some works aiming at unseen relations.
For example, ZSGAN \cite{qin2020generative} and OntoZSL \cite{geng2021ontozsl} leverage generative adversarial networks (GANs) to synthesize valid inductive embeddings for unseen relations conditioned on their external resources such as textual descriptions and ontological schemas.
However, they are also partially inductive with the constraints that each entity that appears in the testing set has appeared during training (i.e., all the entities are seen).

\subsubsection{Fully Inductive}\label{related_work_fully_inductive}
The fully inductive KGC settings aim to address unseen entities and unseen relations simultaneously. 
Such settings are quite common in involving KGs but are more challenging and relatively underexplored.
One of them is a simple extension of the text-based partially inductive method, which also provides text resources for unseen relations as for unseen entities \cite{wang2021structure}.
However, the extra computation cost on massive text resources, especially using large-scale pre-trained language models, is actually a serious turn-off.

Another work that is more related to ours is MaKEr \cite{MaKEr}, a method that also depends on the local graph structure to deal with the unseen entities and unseen relations. 
In comparison to the meta learning based framework used in MaKEr, in our paper, we instead perform reasoning over the enclosing subgraph of a candidate triple using a message passing network.
Moreover, MaKEr actually requires that a part of the relations in the local neighborhood to be seen, and only supports the setting of \textit{testing with semi unseen relations}.
In contrast, our method can support this setting as well as another more challenging setting of \textit{testing with fully unseen relations} where all the relations in the local subgraph are unseen.

\section{Conclusion and Outlook}
In this paper, we propose RMPI, a
relational message passing network for subgraph-based inductive reasoning, which focuses on the relations in the local subgraph and directly passes \revise{messages} from relations to relations iteratively to infer missing triples in a KG.
%the target relation between a pair of entities.
It supports not only partially inductive KGC with only unseen entities but also the  more challenging problem of fully inductive KGC. %involving unseen entities and unseen relations simultaneously.
To implement RMPI, we \textit{(i)} developed an effective graph pruning strategy
to improve the reasoning efficiency;
\textit{(ii)} applied a target relation-aware neighborhood attention mechanism to better aggregate features from the neighborhood with edge types of the relational graph utilized;
%we leverage diverse types of edges in the relational graph and apply a target relation-aware neighborhood attention;
%
\textit{(iii)} utilized discriminative features from the surrounding disclosing subgraphs to deal with empty enclosing subgraphs;
%around the target triple, we propose to explore discriminative features from the surrounding disclosing subgraph so as to score triple more accurately;
and \textit{(iv)} injected the KG's ontological schema for richer relation semantics.
In evaluation, we have performed extensive experiments on many different benchmarks with baselines of state-of-the-art or typical partially inductive and fully inductive KGC methods including TACT, MaKEr, GraIL, CoMPILE and so on.
RMPI often achieves better performance than the baselines, and the effectiveness of its technical components, such as the injection of ontological schema, the target relation-aware neighborhood attention mechanism, and the aggregation on disclosing subgraphs, has been fully verified.

In the future, we plan to study the following aspects:
1) developing more robust solutions to integrate the ontological schema for inductive KGC; 
2) assembling nonnegligible reasoning clues from entities for relational message passing;
and 3) performing more comprehensive evaluations on fully inductive KGC with both seen and unseen entities.
%where the prediction involving seen entities are also tested.

% \begin{table}[htbp]
% \caption{Table Type Styles}
% \begin{center}
% \begin{tabular}{|c|c|c|c|}
% \hline
% \textbf{Table}&\multicolumn{3}{|c|}{\textbf{Table Column Head}} \\
% \cline{2-4} 
% \textbf{Head} & \textbf{\textit{Table column subhead}}& \textbf{\textit{Subhead}}& \textbf{\textit{Subhead}} \\
% \hline
% copy& More table copy$^{\mathrm{a}}$& &  \\
% \hline
% \multicolumn{4}{l}{$^{\mathrm{a}}$Sample of a Table footnote.}
% \end{tabular}
% \label{tab1}
% \end{center}
% \end{table}

\section*{Acknowledgment}

This work is partially funded by NSFCU19B2027/91846204, Zhejiang Provincial Natural Science Foundation of China (No. Q23F020051), 
%Jiaoyan Chen is founded by
the EPSRC project ConCur (EP/V050869/1) and %. Jeff Z. Pan is partially supported by
the Chang Jiang Scholars Program (J2019032).

% \section*{References}

% \todo{Unless there are six authors or more give all authors' names; do not use 
% ``et al.''. Papers that have not been published, even if they have been 
% submitted for publication, should be cited as ``unpublished'' \cite{b4}. Papers 
% that have been accepted for publication should be cited as ``in press'' \cite{b5}. }

\bibliographystyle{plain}
\bibliography{reference}

\begin{thebibliography}{10}

\bibitem{ali2021improving}
Mehdi Ali, Max Berrendorf, Mikhail Galkin, Veronika Thost, Tengfei Ma, Volker
  Tresp, and Jens Lehmann.
\newblock Improving inductive link prediction using hyper-relational facts.
\newblock In {\em ISWC}, pages 74--92, 2021.

\bibitem{ARWP2021}
Hiba Arnaout, Simon Razniewski, Gerhard Weikum, and Jeff~Z. Pan.
\newblock Negative statements considered useful.
\newblock {\em ournal of Web Semantics}, 71, 2021.

\bibitem{auer2007dbpedia}
S{\"o}ren Auer, Christian Bizer, Georgi Kobilarov, Jens Lehmann, Richard
  Cyganiak, and Zachary Ives.
\newblock {DBpedia}: A nucleus for a web of open data.
\newblock In {\em The semantic web}, pages 722--735. Springer, 2007.

\bibitem{baek2020learning}
Jinheon Baek, Dong~Bok Lee, and Sung~Ju Hwang.
\newblock Learning to extrapolate knowledge: Transductive few-shot out-of-graph
  link prediction.
\newblock {\em Advances in Neural Information Processing Systems}, 33:546--560,
  2020.

\bibitem{bhowmik2020explainable}
Rajarshi Bhowmik and Gerard de~Melo.
\newblock Explainable link prediction for emerging entities in knowledge
  graphs.
\newblock In {\em {ISWC} {(1)}}, volume 12506 of {\em Lecture Notes in Computer
  Science}, pages 39--55. Springer, 2020.

\bibitem{transe}
Antoine Bordes, Nicolas Usunier, Alberto Garcia-Duran, Jason Weston, and Oksana
  Yakhnenko.
\newblock Translating embeddings for modeling multi-relational data.
\newblock {\em Advances in Neural Information Processing Systems}, 26, 2013.

\bibitem{TACT}
Jiajun Chen, Huarui He, Feng Wu, and Jie Wang.
\newblock Topology-aware correlations between relations for inductive link
  prediction in knowledge graphs.
\newblock In {\em Proceedings of the AAAI Conference on Artificial
  Intelligence}, volume~35, pages 6271--6278, 2021.

\bibitem{chen2021knowledge}
Jiaoyan Chen, Yuxia Geng, Zhuo Chen, Ian Horrocks, Jeff~Z Pan, and Huajun Chen.
\newblock Knowledge-aware zero-shot learning: Survey and perspective.
\newblock In {\em IJCAI Survey Track}, 2021.

\bibitem{chen2021low}
Jiaoyan Chen, Yuxia Geng, Zhuo Chen, Jeff~Z Pan, Yuan He, Wen Zhang, Ian
  Horrocks, and Huajun Chen.
\newblock Low-resource learning with knowledge graphs: A comprehensive survey.
\newblock {\em unpublished}.

\bibitem{MaKEr}
Mingyang Chen, Wen Zhang, Zhen Yao, Xiangnan Chen, Mengxiao Ding, Fei Huang,
  and Huajun Chen.
\newblock Meta-learning based knowledge extrapolation for knowledge graphs in
  the federated setting.
\newblock In {\em IJCAI}, 2022.

\bibitem{chen2020knowledge}
Zhe Chen, Yuehan Wang, Bin Zhao, Jing Cheng, Xin Zhao, and Zongtao Duan.
\newblock Knowledge graph completion: A review.
\newblock {\em IEEE Access}, 8:192435--192456, 2020.

\bibitem{WN18RR}
Tim Dettmers, Pasquale Minervini, Pontus Stenetorp, and Sebastian Riedel.
\newblock Convolutional 2d knowledge graph embeddings.
\newblock In {\em Proceedings of the AAAI conference on artificial
  intelligence}, volume~32, 2018.

\bibitem{ConvE}
Tim Dettmers, Pasquale Minervini, Pontus Stenetorp, and Sebastian Riedel.
\newblock Convolutional 2d knowledge graph embeddings.
\newblock In {\em AAAI}, 2018.

\bibitem{farber2018linked}
Michael F{\"a}rber, Frederic Bartscherer, Carsten Menne, and Achim Rettinger.
\newblock Linked data quality of dbpedia, freebase, opencyc, wikidata, and
  yago.
\newblock {\em Semantic Web}, 9(1):77--129, 2018.

\bibitem{geng2021ontozsl}
Yuxia Geng, Jiaoyan Chen, Zhuo Chen, Jeff~Z Pan, Zhiquan Ye, Zonggang Yuan,
  Yantao Jia, and Huajun Chen.
\newblock Ontozsl: Ontology-enhanced zero-shot learning.
\newblock In {\em WebConf}, pages 3325--3336, 2021.

\bibitem{geng2022benchmarking}
Yuxia Geng, Jiaoyan Chen, Xiang Zhuang, Zhuo Chen, Jeff~Z Pan, Juan Li,
  Zonggang Yuan, and Huajun Chen.
\newblock Benchmarking knowledge-driven zero-shot learning.
\newblock {\em Journal of Web Semantics}, 2022.

\bibitem{hamaguchi2017knowledge}
Takuo Hamaguchi, Hidekazu Oiwa, Masashi Shimbo, and Yuji Matsumoto.
\newblock Knowledge transfer for out-of-knowledge-base entities : {A} graph
  neural network approach.
\newblock In {\em {IJCAI}}, pages 1802--1808, 2017.

\bibitem{GraphSAGE}
Will Hamilton, Zhitao Ying, and Jure Leskovec.
\newblock Inductive representation learning on large graphs.
\newblock {\em Advances in Neural Information Processing Systems}, 30, 2017.

\bibitem{harary1960some}
Frank Harary and Robert~Z Norman.
\newblock Some properties of line digraphs.
\newblock {\em Rendiconti del circolo matematico di palermo}, 9(2):161--168,
  1960.

\bibitem{CoMPILE}
Sijie Mai, Shuangjia Zheng, Yuedong Yang, and Haifeng Hu.
\newblock Communicative message passing for inductive relation reasoning.
\newblock In {\em AAAI}, pages 4294--4302, 2021.

\bibitem{meilicke2018fine}
Christian Meilicke, Manuel Fink, Yanjie Wang, Daniel Ruffinelli, Rainer
  Gemulla, and Heiner Stuckenschmidt.
\newblock Fine-grained evaluation of rule-and embedding-based systems for
  knowledge graph completion.
\newblock In {\em International Semantic Web Conference}, pages 3--20.
  Springer, 2018.

\bibitem{RuleN}
Christian Meilicke, Manuel Fink, Yanjie Wang, Daniel Ruffinelli, Rainer
  Gemulla, and Heiner Stuckenschmidt.
\newblock Fine-grained evaluation of rule-and embedding-based systems for
  knowledge graph completion.
\newblock In {\em International semantic web conference}, pages 3--20.
  Springer, 2018.

\bibitem{mitchell2018never}
Tom Mitchell, William Cohen, Estevam Hruschka, Partha Talukdar, Bishan Yang,
  Justin Betteridge, Andrew Carlson, Bhavana Dalvi, Matt Gardner, Bryan Kisiel,
  et~al.
\newblock Never-ending learning.
\newblock {\em Communications of the ACM}, 61(5):103--115, 2018.

\bibitem{Pan2009}
Jeff~Z. Pan.
\newblock Resource description framework.
\newblock In {\em Handbook on Ontologies}, pages 71--90. Springer, 2009.

\bibitem{PVGW2017}
Jeff~Z. Pan, Guido Vetere, Jose~Manue Gomez-Perez, and Honghan Wu.
\newblock {\em Exploiting Linked Data and Knowledge Graphs in Large
  Organisations}.
\newblock Springer, 2017.

\bibitem{qin2020generative}
Pengda Qin, Xin Wang, Wenhu Chen, Chunyun Zhang, Weiran Xu, and William~Yang
  Wang.
\newblock Generative adversarial zero-shot relational learning for knowledge
  graphs.
\newblock In {\em Proceedings of the AAAI Conference on Artificial
  Intelligence}, volume~34, pages 8673--8680, 2020.

\bibitem{DRUM}
Ali Sadeghian, Mohammadreza Armandpour, Patrick Ding, and Daisy~Zhe Wang.
\newblock Drum: End-to-end differentiable rule mining on knowledge graphs.
\newblock {\em Advances in Neural Information Processing Systems}, 32, 2019.

\bibitem{RGCN}
Michael~Sejr Schlichtkrull, Thomas~N. Kipf, Peter Bloem, Rianne van~den Berg,
  Ivan Titov, and Max Welling.
\newblock Modeling relational data with graph convolutional networks.
\newblock In {\em {ESWC}}, pages 593--607, 2018.

\bibitem{shah2019open}
Haseeb Shah, Johannes Villmow, Adrian Ulges, Ulrich Schwanecke, and Faisal
  Shafait.
\newblock An open-world extension to knowledge graph completion models.
\newblock In {\em AAAI}, pages 3044--3051, 2019.

\bibitem{rotate}
Zhiqing Sun, Zhi{-}Hong Deng, Jian{-}Yun Nie, and Jian Tang.
\newblock {RotatE}: Knowledge graph embedding by relational rotation in complex
  space.
\newblock In {\em {ICLR} (Poster)}. OpenReview.net, 2019.

\bibitem{GraIL}
Komal Teru, Etienne Denis, and Will Hamilton.
\newblock Inductive relation prediction by subgraph reasoning.
\newblock In {\em International Conference on Machine Learning}, pages
  9448--9457. PMLR, 2020.

\bibitem{FB15k237}
Kristina Toutanova, Danqi Chen, Patrick Pantel, Hoifung Poon, Pallavi
  Choudhury, and Michael Gamon.
\newblock Representing text for joint embedding of text and knowledge bases.
\newblock In {\em EMNLP}, pages 1499--1509, 2015.

\bibitem{complex}
Th{\'e}o Trouillon, Johannes Welbl, Sebastian Riedel, {\'E}ric Gaussier, and
  Guillaume Bouchard.
\newblock Complex embeddings for simple link prediction.
\newblock In {\em International Conference on Machine Learning}, pages
  2071--2080. PMLR, 2016.

\bibitem{CompGCN}
Shikhar Vashishth, Soumya Sanyal, Vikram Nitin, and Partha~P. Talukdar.
\newblock Composition-based multi-relational graph convolutional networks.
\newblock In {\em {ICLR}}. OpenReview.net, 2020.

\bibitem{vrandevcic2014wikidata}
Denny Vrande{\v{c}}i{\'c} and Markus Kr{\"o}tzsch.
\newblock Wikidata: a free collaborative knowledgebase.
\newblock {\em Communications of the ACM}, 57(10):78--85, 2014.

\bibitem{wang2021structure}
Bo~Wang, Tao Shen, Guodong Long, Tianyi Zhou, Ying Wang, and Yi~Chang.
\newblock Structure-augmented text representation learning for efficient
  knowledge graph completion.
\newblock In {\em WebConf}, pages 1737--1748, 2021.

\bibitem{PATHCON}
Hongwei Wang, Hongyu Ren, and Jure Leskovec.
\newblock Relational message passing for knowledge graph completion.
\newblock In {\em SIGKDD}, pages 1697--1707, 2021.

\bibitem{LAN}
Peifeng Wang, Jialong Han, Chenliang Li, and Rong Pan.
\newblock Logic attention based neighborhood aggregation for inductive
  knowledge graph embedding.
\newblock In {\em Proceedings of the AAAI Conference on Artificial
  Intelligence}, volume~33, pages 7152--7159, 2019.

\bibitem{wang2017knowledge}
Quan Wang, Zhendong Mao, Bin Wang, and Li~Guo.
\newblock Knowledge graph embedding: A survey of approaches and applications.
\newblock {\em IEEE Transactions on Knowledge and Data Engineering},
  29(12):2724--2743, 2017.

\bibitem{transh}
Zhen Wang, Jianwen Zhang, Jianlin Feng, and Zheng Chen.
\newblock Knowledge graph embedding by translating on hyperplanes.
\newblock In {\em AAAI}, 2014.

\bibitem{WPKD2020}
Kemas Wiharja, Jeff~Z. Pan, Martin~J. Kollingbaum, and Yu~Deng.
\newblock {Schema Aware Iterative Knowledge Graph Completion}.
\newblock {\em Journal of Web Semantics}, 2020.

\bibitem{xie2016representation}
Ruobing Xie, Zhiyuan Liu, Jia Jia, Huanbo Luan, and Maosong Sun.
\newblock Representation learning of knowledge graphs with entity descriptions.
\newblock In {\em AAAI}, 2016.

\bibitem{IKRL}
Ruobing Xie, Zhiyuan Liu, Huanbo Luan, and Maosong Sun.
\newblock Image-embodied knowledge representation learning.
\newblock In {\em {IJCAI}}, pages 3140--3146, 2017.

\bibitem{NELL995}
Wenhan Xiong, Thien Hoang, and William~Yang Wang.
\newblock Deeppath: {A} reinforcement learning method for knowledge graph
  reasoning.
\newblock In {\em {EMNLP}}, pages 564--573, 2017.

\bibitem{distmult}
Bishan Yang, Wen{-}tau Yih, Xiaodong He, Jianfeng Gao, and Li~Deng.
\newblock Embedding entities and relations for learning and inference in
  knowledge bases.
\newblock In {\em {ICLR} (Poster)}, 2015.

\bibitem{NeuralLP}
Fan Yang, Zhilin Yang, and William~W Cohen.
\newblock Differentiable learning of logical rules for knowledge base
  reasoning.
\newblock {\em Advances in Neural Information Processing Systems}, 30, 2017.

\bibitem{kgbert}
Liang Yao, Chengsheng Mao, and Yuan Luo.
\newblock {KG-BERT}: {BERT} for knowledge graph completion.
\newblock {\em unpublished}.

\end{thebibliography}
% \bibitem{b1} G. Eason, B. Noble, and I. N. Sneddon, ``On certain integrals of Lipschitz-Hankel type involving products of Bessel functions,'' Phil. Trans. Roy. Soc. London, vol. A247, pp. 529--551, April 1955.
% \bibitem{b2} J. Clerk Maxwell, A Treatise on Electricity and Magnetism, 3rd ed., vol. 2. Oxford: Clarendon, 1892, pp.68--73.
% \bibitem{b3} I. S. Jacobs and C. P. Bean, ``Fine particles, thin films and exchange anisotropy,'' in Magnetism, vol. III, G. T. Rado and H. Suhl, Eds. New York: Academic, 1963, pp. 271--350.
% \bibitem{b4} K. Elissa, ``Title of paper if known,'' unpublished.
% \bibitem{b5} R. Nicole, ``Title of paper with only first word capitalized,'' J. Name Stand. Abbrev., in press.
% \bibitem{b6} Y. Yorozu, M. Hirano, K. Oka, and Y. Tagawa, ``Electron spectroscopy studies on magneto-optical media and plastic substrate interface,'' IEEE Transl. J. Magn. Japan, vol. 2, pp. 740--741, August 1987 [Digests 9th Annual Conf. Magnetics Japan, p. 301, 1982].
% \bibitem{b7} M. Young, The Technical Writer's Handbook. Mill Valley, CA: University Science, 1989.
% \end{thebibliography}

\end{document}